\documentclass[twoside]{article}

\usepackage[accepted]{aistats2022}
\usepackage{color}

\usepackage[round]{natbib}
\usepackage{microtype}
\usepackage{graphicx}
\usepackage{subfigure}
\usepackage{graphicx}
\usepackage{bm}
\usepackage{amsmath, amssymb}
\usepackage{amsthm}
\usepackage{breqn}
\usepackage[shortlabels]{enumitem}

\theoremstyle{definition}

\usepackage{booktabs} 
\usepackage{siunitx}
\usepackage{listings}
\usepackage{subfigure}
\usepackage{comment}
\usepackage{url}
\usepackage[ruled,vlined]{algorithm2e}
\usepackage{algorithmic}
\usepackage{here}
\usepackage{comment}

\newcommand{\argmin}{\mathop{\rm argmin}\limits}

\begin{document}

%

%
\twocolumn[

\aistatstitle{Fixed Support Tree-Sliced Wasserstein Barycenter}

\aistatsauthor{ Yuki Takezawa$^{1,2}$ \And Ryoma Sato$^{1,2}$ \And Zornitsa Kozareva$^3$ \And Sujith Ravi$^4$ \And  Makoto Yamada$^{1,2}$ }

\aistatsaddress{ $^1$Kyoto University \And $^2$RIKEN AIP \And  $^3$Facebook AI Research \And $^4$SliceX AI } ]

\begin{abstract}
The Wasserstein barycenter has been widely studied in various fields, including natural language processing, and computer vision.
However, it requires a high computational cost to solve the Wasserstein barycenter problem 
because the computation of the Wasserstein distance requires a quadratic time with respect to the number of supports. 
By contrast, the Wasserstein distance on a tree, called the tree-Wasserstein distance, can be computed in linear time and allows for the fast comparison of a large number of distributions.
In this study, we propose a barycenter under the tree-Wasserstein distance, called the fixed support tree-Wasserstein barycenter (FS-TWB) 
and its extension, called the fixed support tree-sliced Wasserstein barycenter (FS-TSWB). 
More specifically, we first show that the FS-TWB and FS-TSWB problems are convex optimization problems
and can be solved by using the projected subgradient descent.
Moreover, we propose a more efficient algorithm to compute the subgradient and objective function value by using the properties of tree-Wasserstein barycenter problems.
Through real-world experiments, we show that, by using the proposed algorithm, the FS-TWB and FS-TSWB can be solved two orders of magnitude faster than the original Wasserstein barycenter.
\end{abstract}

\section{Introduction}
To measure the dissimilarity between distributions, the Wasserstein distance is widely used.
The Wasserstein distance can be solved by using linear programming.
However, its time complexity is cubic with respect to the number of supports. 
\citet{cuturi2013sinkhorn} proposed the entropic regularized Wasserstein distance,
which can be computed using the matrix scaling algorithm in quadratic time with respect to the number of supports.
Following this work,
the Wasserstein distance has been applied in many fields such as
document classification \citep{kusner2015from,hung2016supervised} and generative models \citep{arjovsky2017wasserstein}, among other areas.

One of the fundamental topics related to the Wasserstein distance is the Wasserstein barycenter problem,
which has been applied to many applications such as natural language processing \citep{xu2018distilled},
image processing \citep{simon2020barycenter,julien2011wasserstein}, and so on \citep{dognin2018wasserstein,solomon2015convolutional}.
Based on the entropic regularized Wasserstein distance,
\citet{benamou2015iterative} showed that the entropic regularized Wasserstein barycenter problem can be solved using the iterative Bregman projection. 
Many researchers have recently tried to further reduce the computational cost of the Wasserstein barycenter problem
\citep{claici2018stochastic, ge2019interior, lin2020fixed, guminov2021accelerated, dvinskikh2021improved}.

\begin{figure}[t!]
\vskip -0.10 in
  \begin{minipage}[b]{.5\columnwidth}
    \centering
    \includegraphics[width=.65\hsize]{{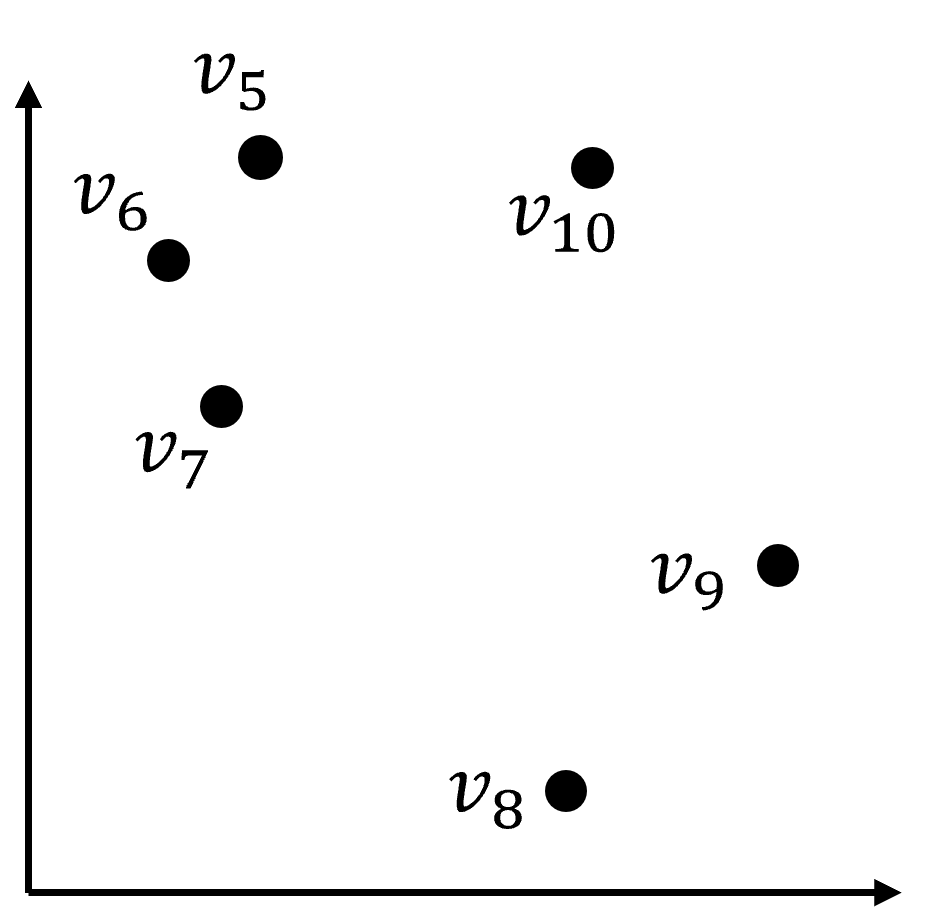}}%
  \end{minipage}%
  \begin{minipage}[b]{.5\columnwidth}
    \centering
    \includegraphics[width=.65\hsize]{{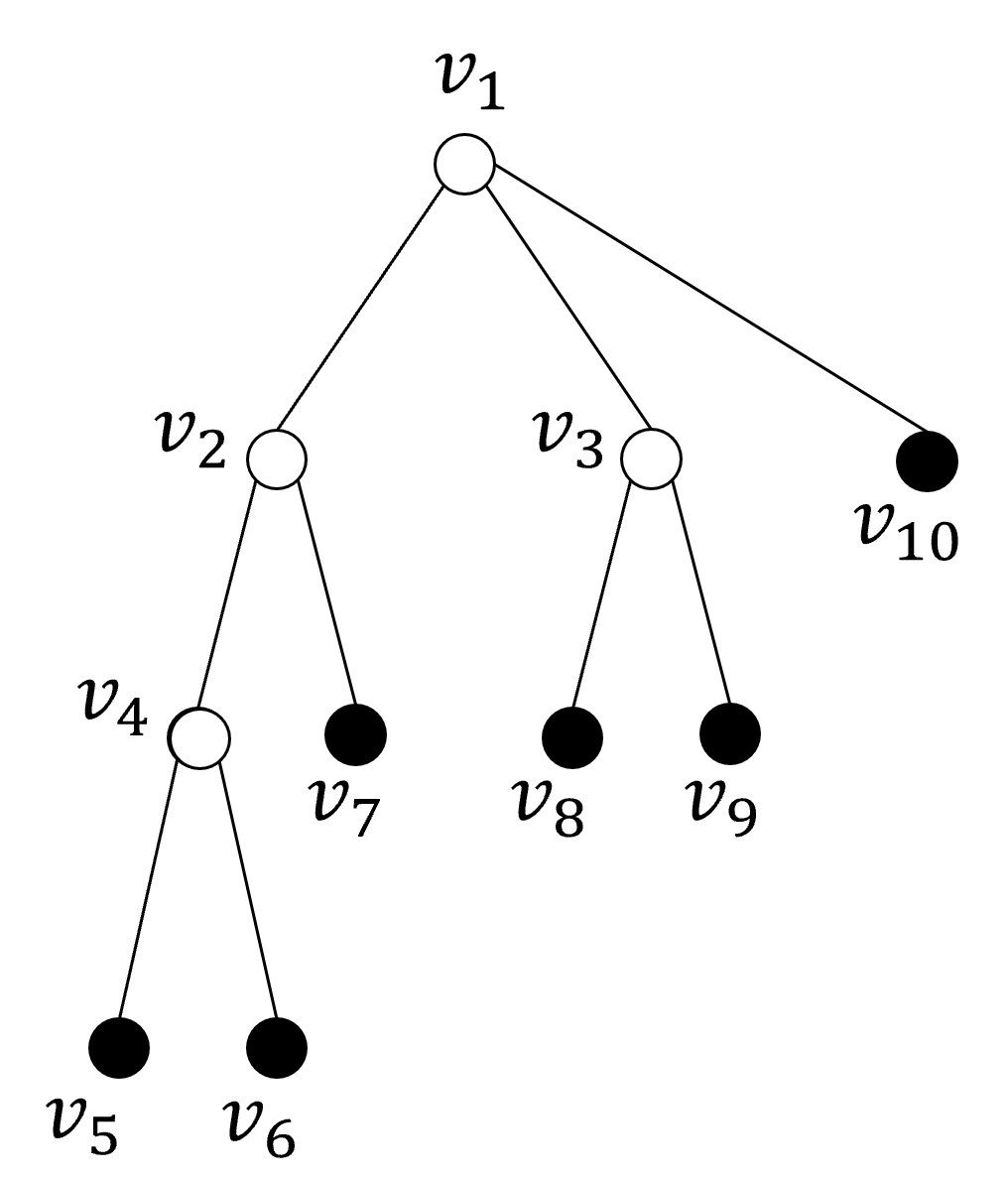}}
  \end{minipage}
\vspace{-0.3in}
\caption{Illustration of the original space (left) and tree (right). A black node has a corresponding element in the original space, but a white node has no corresponding element.}
\label{fig:illustration_tree}
\vspace{-0.1 in}
\end{figure}

However, the Wasserstein barycenter still suffers from a high computational cost because the computation of the Wasserstein distance itself is expensive.
To accelerate the computation of the Wasserstein distance, various techniques have been proposed, such as the sliced Wasserstein distance \citep{julien2011wasserstein,kolouri2018sliced,kolouri2019generalized,deshpande2019max}, 
its generalization, the tree-Wasserstein distance \citep{indyk2003fast,le2019tree,backurs2020scalable,sato2020fast,le2021entropy,takezawa2021supervised},
and other versions \citep{tong2021diffusion}.
The key advantage of the tree-Wasserstein distance is that it has a closed-form solution, which can be computed in linear time with respect to the number of nodes.
Recently, utilizing this advantage, \citet{le2020treewasserstein} studied a barycenter problem under the tree-Wasserstein distance,
and showed that the tree-Wasserstein barycenter problem can be solved faster than the Wasserstein barycenter problem.
They showed that their proposed tree-Wasserstein barycenter works well experimentally. 
However, their barycenter problem is not a proper barycenter problem on a tree. 
Fig. \ref{fig:illustration_tree} shows an illustration of a tree used for the tree-Wasserstein distance.
In general, for the tree-Wasserstein distance, we assign the probability only to the black nodes of a tree. 
Howerver, \citet{le2020treewasserstein} assumes to have probability on all nodes.
This violates the assumption of the tree-Wasserstein distance.  

In this study, we properly formulate the barycenter problem under the tree-Wasserstein distance 
and propose an efficient optimization algorithm.
More specifically, we constrain a barycenter to have the probability on only black nodes in Fig. \ref{fig:illustration_tree}, 
and then employ a matrix-form formulation of the tree-Wasserstein distance \citep{takezawa2021supervised}.
This formulation results in a convex optimization problem.
We refer to this single-tree version of the tree-Wasserstein barycenter problem as the \textit{fixed support tree-Wasserstein barycenter} (FS-TWB) problem. 
Moreover, we propose the \textit{fixed support tree-sliced Wasserstein barycenter} (FS-TSWB) problem,
which is the barycenter under the tree-sliced Wasserstein distance (i.e., multiple trees) \citep{le2019tree}.
We then propose a more efficient algorithm to compute the subgradient and objective function value by using the properties of the FS-TWB and FS-TSWB problems.
Through experiments on real large-scale data, 
we show that the FS-TWB and FS-TSWB problems can be solved two orders of magnitude faster than the original Wasserstein barycenter problem. 
Moreover, by sampling multiple trees, we show that the original Wasserstein barycenter can be efficiently approximated using the FS-TSWB.

\textbf{Notation:}
We denote $[\![ n ]\!] = \{1, 2, \ldots, n\}$ for any $n \in \mathbb{N}$.
$[\mathbf{a}]_i$ denotes an $i$-th element of the vector $\mathbf{a}$.
$\mathbf{I}$ is the identity matrix.
$\mathbf{1}_n$ is an $n$-dimensional vector with all ones,
and $\mathbf{0}_n$ is an $n$-dimensional vector with all zeros.
\section{Related Work}
\subsection{Wasserstein Distance}
Let $P(\Omega)$ be the set of Borel probability measures on $\Omega$. 
Let $d:\Omega \times \Omega \rightarrow \mathbb{R}_{+}$ be a metric.
Given two probability measures $\mu_i, \mu_j \in P(\Omega)$,
the Wasserstein distance is defined as follows:
\begin{align*}
    W_d (\mu_i, \mu_j) = \inf_{\gamma \in \Pi(\mu_i, \mu_j)} \int_{\Omega \times \Omega} d(x, y) \gamma(dx, dy),
\end{align*}
where $\Pi(\mu_i, \mu_j)$ is the set of couplings between $\mu_i$ and $\mu_j$.
The Wasserstein distance can be computed by linear programming.
However, linear programming requires cubic time with respect to the number of supports.
To reduce this time complexity, \citet{cuturi2013sinkhorn} proposed adding entropic regularization to the Wasserstein distance,
which can be computed using the Sinkhorn algorithm in quadratic time.
In some special cases, the Wasserstein distance has a closed-form solution.
For example, if $\Omega$ is a one-dimensional space, 
the Wasserstein distance can be computed using the sorting algorithm.
Using this property, the sliced Wasserstein distance has been proposed \citep{julien2011wasserstein,kolouri2018sliced,kolouri2019generalized,deshpande2019max}. 
In the next section, we introduce the case in which the metric $d$ is a tree metric.

\subsection{Tree-Wasserstein Distance}
When $d$ is a tree metric,
the Wasserstein distance is called the tree-Wasserstein distance.
Let $\mathcal{T} = (\bm{V}, \bm{E})$ be a tree with $v_1$ as the root.
For any node $v \in \bm{V} \setminus \{v_1\}$,
let $w_v$ be the length of the edge between $v$ and its parent node.
For the simplicity, we define $w_{v_1}=0$. 
Let $d_{\mathcal{T}}:\bm{V} \times \bm{V} \rightarrow \mathbb{R}_{+}$ be the total length of the path between two nodes.
Given two probability measures $\mu_i$, $\mu_j \in P(\bm{V})$,
the tree-Wasserstein distance can be computed as follows:
\begin{align}
    W_{d_{\mathcal{T}}} (\mu_i, \mu_j) = \sum_{v \in \bm{V}} w_v | \mu_i(\Gamma(v)) - \mu_j(\Gamma(v)) |,
\end{align}
where $\Gamma(v)$ denotes the set of nodes contained in the subtree rooted at $v$ \citep{le2019tree}.
Note that, because a chain is a tree, 
the tree-Wasserstein distance is considered as a generalization of the Wasserstein distance on a one-dimensional space.
The key of the tree-Wasserstein distance is that it has the closed-form solution,
which can be computed in linear time with respect to the number of nodes.

To compute the tree-Wasserstein distance,
we need to build the tree metric.
For embedding the coordinates in the original space $\Omega$ into a tree,
Quadtree \citep{indyk2003fast} and a clustering-based method \citep{le2019tree} have been proposed.
Fig. \ref{fig:illustration_tree} shows an illustration of the original space and the tree.
In a tree constructed using these methods, nodes are classified into two groups: 
\textit{leaf nodes} and \textit{internal nodes} \citep{takezawa2021supervised}.
A leaf node corresponds to an element in $\Omega$,
and an internal node does not correspond to any element in $\Omega$.
In Fig. \ref{fig:illustration_tree}, black nodes are leaf nodes, and white nodes are internal nodes.
We denote $\bm{V}_{\text{leaf}}$ as the set of leaf nodes
and $\bm{V}_{\text{in}} = \bm{V} \setminus \bm{V}_{\text{leaf}}$ as the set of internal nodes.
(i.e., $\bm{V}_{\text{leaf}} = \Omega$).
In general, the given probability measures to be compared by the tree-Wasserstein distance satisfies $\mu(\bm{V}_{\text{in}})=0$.

Recently, \citet{takezawa2021supervised} showed the matrix-form formulation of the tree-Wasserstein distance.
Let $\mathcal{T}^{\prime} = (\bm{V}, \bm{E}^{\prime})$ be the directed tree with $v_1$ as the root,
which has directed edges from $v \in \bm{V} \setminus \{v_1\}$ to its parent node in $\mathcal{T}$.
We denote $\bm{V}_{\text{in}} = \{ v_1, v_2, \ldots, v_{|\bm{V}_{\text{in}}|}\}$ 
and $\bm{V}_{\text{leaf}} = \{ v_{|\bm{V}_{\text{in}}| + 1}, v_{|\bm{V}_{\text{in}}| + 2}, \ldots, v_{|\bm{V}|} \}$.
Without a lack of generality, we assume $i>j$ for all edges $(v_i, v_j) \in \bm{E}^\prime$.
Let $\mathbf{D}_{\text{par}}$ be an adjacency matrix of $\mathcal{T}^{\prime}$ and $\mathbf{w}_v = (w_{v_1}, \ldots, w_{v_{|\bm{V}|}})^\top$.
The tree-Wasserstein distance between two probability measures $\mu_i, \mu_j \in P(\bm{V}_{\text{leaf}})$ can be computed as follows:
\begin{align}
    W_{d_{\mathcal{T}}}(\mu_i, \mu_j) \! = \! \left\| \mathbf{w}_v \circ (\mathbf{I} - \mathbf{D}_{\text{par}})^{-1}
    \left( \!
    \begin{array}{c}
    \mathbf{0}_{| \bm{V}_{\text{in}} |} \\
    \mathbf{a}_i \! - \! \mathbf{a}_j
    \end{array} \! \right) \right\|_1,
\label{eq:stw}
\end{align}
where $\circ$ denotes the element-wise Hadamard product, $\mathbf{a}_i$ and $\mathbf{a}_j$ are $|\bm{V}_{\text{leaf}}|$-dimensional vectors whose $k$-th elements are $\mu_i(v_{|\bm{V}_{\text{in}}| + k})$ and $\mu_j(v_{|\bm{V}_{\text{in}}| + k})$ respectively,
and $\mathbf{0}_{| \bm{V}_{\text{in}} |}$ means that $\mu_i(\bm{V}_{\text{in}})=0$ and $\mu_j(\bm{V}_{\text{in}})=0$.
Considering that leaf nodes have no child nodes,
$\mathbf{D}_{\text{par}}$ is partitioned into four blocks as follows:
\begin{align*}
    \mathbf{D}_{\text{par}} = \begin{pmatrix}
    \mathbf{D}_1 & \mathbf{D}_2 \\
    \mathbf{0} & \mathbf{0}
    \end{pmatrix} ,
\end{align*}
where $\mathbf{D}_1$ is a $|\bm{V}_{\text{in}}| \times |\bm{V}_{\text{in}}|$ matrix,
which is the adjacency matrix of the tree consisting of the internal nodes, 
and $\mathbf{D}_2$ is a $|\bm{V}_{\text{in}}| \times |\bm{V}_{\text{leaf}}|$ matrix.
The inverse matrix is then computed as follows:
\begin{align*}
    (\mathbf{I} - \mathbf{D}_{\text{par}})^{-1} = 
    \begin{pmatrix}
    (\mathbf{I} - \mathbf{D}_1)^{-1} & (\mathbf{I} - \mathbf{D}_1)^{-1}\mathbf{D}_2 \\
    \bm{0} & \mathbf{I}
    \end{pmatrix}.
\end{align*}
In other words,
if $v_j \in \Gamma(v_i)$,
$[(\mathbf{I} - \mathbf{D}_{\text{par}})^{-1}]_{i j}$ is one, and is zero otherwise.
Let $D$ be the depth of the tree $\mathcal{T}$.
Because $w_{v_1}=0$ and $|\{ u | v \in \Gamma(u)\} \setminus \{ v_1 \}|\leq D$ for all $v \in \bm{V}$, 
$\mathbf{w}_v\circ (\mathbf{I}-\mathbf{D}_{\text{par}})^{-1}$ is a sparse matrix whose each column has at most $D$ non-zero elements.

\subsection{Wasserstein Barycenter}
Given a set of probability measures $\{ \mu_i | \mu_i \in P(\Omega) \}_{i=1}^{N}$, 
the Wasserstein barycenter is defined as follows:
\begin{align}
\label{eq:fs_wb}
    \overline{\mu} \in \argmin_{\mu \in P(\Omega)} \frac{1}{N} \left( \sum_{i=1}^{N} W_d (\mu, \mu_i) \right).
\end{align}
When the set of supports is fixed, the Wasserstein barycenter is called the fixed support Wasserstein barycenter (FS-WB);
otherwise, it is called the free support Wasserstein barycenter. 
In this study, we consider the case in which the set of supports is fixed.
However, even if the set of supports is fixed, it is intractable to solve the FS-WB problem exactly.
Following the previous work \citep{cuturi2013sinkhorn},
\citet{cuturi2014fast} showed that 
the barycenter under the entropic regularized Wasserstein distance can be efficiently solved.
\citet{benamou2015iterative} showed that the barycenter under the entropic regularized Wasserstein distance can be solved using the iterative Bregman projection (IBP).
However, the time complexity of the IBP is $O(N |\Omega|^2)$ \citep{kroshnin2019on},
and it still requires a high computational cost to solve the FS-WB problem.

Utilizing the property in which the sliced Wasserstein distance has a closed form solution, 
\citet{julien2011wasserstein} and \citet{bonneel2015sliced} studied the sliced Wasserstein barycenter.
Recently, \citet{le2020treewasserstein} proposed the tree-Wasserstein barycenter on $\bm{V}$, 
and showed that the tree-Wasserstein barycenter can be computed faster than the FS-WB.
Given a set of probability measures $\{ \mu_i | \mu_i \in P(\bm{V}_{\text{leaf}}) \}_{i=1}^{N}$,
the tree-Wasserstein barycenter on $\bm{V}$ is defined as follows:
\begin{align}
\label{eq:free-support-twb}
    \overline{\mu} \in \argmin_{\mu \in P(\bm{V})} \frac{1}{N} \left( \sum_{i=1}^{N} W_{d_{\mathcal{T}}} (\mu, \mu_i) \right).
\end{align}
However, our goal is to compute a barycenter on $\Omega$ fast
by approximating the Wasserstein distance with the tree-Wasserstein distance.
The probability on a leaf node is considered as the probability on the corresponding element in $\Omega$;
however, the probability on an internal node is meaningless
because the internal node has no corresponding elements in $\Omega$.
Therefore, in contrast to this previous work, 
we formulate the tree-Wasserstein barycenter on $\bm{V}_{\text{leaf}}$, called the FS-TWB,
and propose an algorithm to solve it.

\section{Proposed Method}
In this section, 
we first formulate the FS-TWB problem
and propose an efficient algorithm to solve the FS-TWB problem.
We then propose an extension of the FS-TWB problem, called the FS-TSWB problem, and propose an algorithm to solve the FS-TSWB problem.

\subsection{Fixed Support Tree-Wasserstein Barycenter}
Given a set of probability measures $\{ \mu_i | \mu_i \in P(\Omega) \}_{i=1}^{N}$,
our goal is to compute the barycenter on $\Omega$ fast using the tree-Wasserstein distance.
Let $\mathcal{T} = (\bm{V}, \bm{E})$ be a tree 
that is constructed by the Quadtree \citep{indyk2003fast} or the clustering-based method \citep{le2019tree}.
$\bm{V}_{\text{leaf}}$ denotes the set of leaf nodes, $\bm{V}_{\text{in}}$ denotes the set of internal nodes,
and $D$ denotes the depth of the tree $\mathcal{T}$.
The probability measures on $\Omega$ can be considered as the probability measures on $\bm{V}_{\text{leaf}}$.
Then, given a set of probability measures $\{ \mu_i | \mu_i \in P(\bm{V}_{\text{leaf}})\}_{i=1}^{N}$,
the tree-Wasserstein barycenter on $\bm{V}_{\text{leaf}}$ is defined as follows:
\begin{align}
\label{eq:fs-twb}
    \overline{\mu}_{d_{\mathcal{T}}} \in \argmin_{\mu \in P(\bm{V}_{\text{leaf}})} \frac{1}{N} \left( \sum_{i=1}^{N} W_{d_{\mathcal{T}}} (\mu, \mu_i) \right),
\end{align}
which we refer to as the \textit{fixed support tree-Wasserstein barycenter} (FS-TWB).
In the FS-TWB problem,
we only need to consider the probability measures on $\bm{V}_{\text{leaf}}$.
Combining Eq. \eqref{eq:fs-twb} with Eq. \eqref{eq:stw},
the objective function is rewritten as follows:
\begin{align}
    \mathbf{B} &= 
    \mathbf{w}_v
    \circ
    \begin{pmatrix}
    (\mathbf{I} - \mathbf{D}_1)^{-1}\mathbf{D}_2 \\
    \mathbf{I}
    \end{pmatrix}, \\
    f(\mathbf{a}) &= \frac{1}{N} \sum_{i=1}^{N} \| \mathbf{B} \mathbf{a} - \mathbf{B} \mathbf{a}_i \|_1,
    \label{eq:obj}
\end{align}
where $[\mathbf{a}_i]_k = \mu_i (v_{|\bm{V}_{\text{in}}| + k})$ and $[\mathbf{a}]_k = \mu (v_{|\bm{V}_{\text{in}}| + k})$.
We define $\bm{A} = \{\mathbf{a} \in \mathbb{R}_{+}^{|\bm{V}_{\text{leaf}}|} \mid \|\mathbf{a}\|_1 = 1\}$.
The FS-TWB problem can then be formulated as follows:
\begin{align}
\label{eq:tw_barycenter}
    \overline{\mathbf{a}} \in \argmin_{\mathbf{a} \in \bm{A}} f(\mathbf{a}).
\end{align}

\subsection{Optimization Method}
\label{sec:psd}
The objective function $f$ is a nondifferentiable convex function
and Lipschitz continuous,
and the feasible region $\bm{A}$ is convex.
Therefore, the FS-TWB problem is a convex optimization problem, 
which can be solved by using the projected subgradient descent (PSD) \citep{boyd2003subgradient}.
In other words, the PSD converges to an arbitrarily close approximation to the global minimum value of the FS-TWB problem.
Algorithm \ref{alg:psd} shows the PSD for the FS-TWB problem.
In the following, we describe each modules of this algorithm in detail.

\textbf{Projection onto a simplex.}
The function $\textbf{proj}_{\bm{A}}$ in Algorithm \ref{alg:psd} is the projection of a given vector $\mathbf{x} \in \mathbb{R}^{|\bm{V}_{\text{leaf}}|}$ onto the simplex $\bm{A}$,
which is defined as follows:
\begin{align}
    \text{proj}_{\bm{A}} (\mathbf{x}) = \argmin_{\mathbf{a} \in \bm{A}} \; \|\mathbf{x} - \mathbf{a} \|_2^{2}.
\end{align}
This can be solved using the algorithm proposed by \citet{duchi2008efficient} in $O(|\bm{V_{\text{leaf}}}| \log (|\bm{V_{\text{leaf}}}|))$.

\textbf{Subgradient of $f$.}
One of the subgradients of $f$ at $\mathbf{a}^{(k)}$ is calculated as follows:
\begin{align*}
    \mathbf{g}^{(k)} &= \frac{1}{N} \mathbf{B}^\top \left( \sum_{i=1}^{N} \text{sign} (\mathbf{B} \mathbf{a}^{(k)} - \mathbf{B} \mathbf{a}_i) \right),
\end{align*}
where $\textbf{sign}$ is the element-wise signum function.
Hereafter, we describe the time complexity required to compute $\mathbf{g}^{(k)}$.
For all $i$, $\mathbf{B} \mathbf{a}_i$ needs to be computed only once before starting the iterations.
In addition, $\mathbf{B} \mathbf{a}^{(k)}$ needs to be computed only once per iteration.
Because $\mathbf{B}$ is a sparse matrix that has at most $D |\bm{V}_{\text{leaf}}|$ non-zero elements,
$\mathbf{B} \mathbf{a}^{(k)}$ is computed in $O(D|\bm{V}_{\text{leaf}}|)$.
Therefore, $\mathbf{g}^{(k)}$ is computed in $O(N |\bm{V}| + D |\bm{V}_{\text{leaf}}|)$.
Because the internal nodes that have only one child node can be abbreviated,
we can assume $|\bm{V}| < 2|\bm{V}_{\text{leaf}}|$ without a lack of generality. 
Then, the time complexity required to compute $\mathbf{g}^{(k)}$ is $O((N + D) |\bm{V}_{\text{leaf}}|)$.

\textbf{Objective function value.}
Next, we describe the time complexity required to compute $f(\mathbf{a}^{(k)})$.
Considering that $\mathbf{B} \mathbf{a}^{(k)} - \mathbf{B} \mathbf{a}_i$ is computed when computing the subgradient, 
the time complexity required to compute the objective function value is $O(N|\bm{V}_{\text{leaf}}|)$.
In summary, the time complexity for each iteration of the PSD is $O(|\bm{V}_{\text{leaf}}| (\log(|\bm{V}_{\text{leaf}}|) + N + D))$,
which is faster than the IBP in terms of the number of supports $|\bm{V}_{\text{leaf}}|$.
\begin{algorithm}[tb]
   \caption{PSD for the FS-TWB.}
   \label{alg:psd}
\begin{algorithmic}[1]
   \STATE {\bfseries Input:}  Probability measures $\mathbf{a}_1, \mathbf{a}_2, \ldots, \mathbf{a}_N$, and step size $0 < \gamma_1$ and $0< \gamma_2 \leq 1$.
   \STATE {\bfseries Output:} The FS-TWB.
   \STATE Let $\mathbf{a}^{(0)} \in \bm{A}$. 
   \STATE $\mathbf{a}^{\text{best}} \leftarrow \mathbf{a}^{(0)}$
   \STATE $f^{\text{best}} \leftarrow f(\mathbf{a}^{(0)})$
   \FOR{$k = 0, 1, \ldots, K$}
   \STATE Let $\mathbf{g}^{(k)}$ be an any subgradient of $f$ at $\mathbf{a}^{(k)}$.
   \STATE $\gamma^{(k)} \leftarrow \frac{\gamma_1}{(k + 1)^{\gamma_2} \| \mathbf{g}^{(k)} \|_2}$
   \STATE $\mathbf{a}^{(k + 1)} \leftarrow \text{proj}_{\bm{A}}(\mathbf{a}^{(k)} - \gamma^{(k)} \mathbf{g}^{(k)})$
   \STATE $f^{(k+1)} \leftarrow f(\mathbf{a}^{(k+1)})$
   \IF {$f^{\text{best}} > f^{(k+1)}$}
   \STATE $\mathbf{a}^{\text{best}} \leftarrow \mathbf{a}^{(k+1)}$
   \STATE $f^{\text{best}} \leftarrow f^{(k+1)}$
   \ENDIF
   \ENDFOR
   \STATE {\bfseries return} $\mathbf{a}^{\text{best}}$
\end{algorithmic}
\end{algorithm}
\subsection{Fast Projected Subgradient Descent}
\label{sec:fast_psd}
The bottlenecks of the PSD are two parts: 
the part to compute the subgradient $\mathbf{g}^{(k)}$ 
and the part to compute the objective function value $f(\mathbf{a}^{(k)})$. 
In this section, we propose the algorithm to reduce these time complexity.

\textbf{Subgradient of $f$.}
First, we show the algorithm to reduce the time complexity for computing the subgradient $\mathbf{g}^{(k)}$.
We define $\mathbf{b}^{(k)} = \mathbf{B} \mathbf{a}^{(k)}$, $\mathbf{b}_i =  \mathbf{B} \mathbf{a}_i$ and
$\mathbf{z}^{(k)} = \sum_{i=1}^{N} \text{sign} (\mathbf{b}^{(k)} - \mathbf{b}_i)$. 
(i.e., $\mathbf{g}^{(k)} = \frac{1}{N} \mathbf{B}^\top \mathbf{z}^{(k)}$).
Then, the $j$-th element of $\mathbf{z}^{(k)}$ is computed as follows:
\begin{align}
\label{eq:fast_subgradient}
    [\mathbf{z}^{(k)}]_j =  \sum_{i=1}^{N} \text{sign} \left( [\mathbf{b}^{(k)}]_j - [\mathbf{b}_i]_j \right).
\end{align}
From Eq. \eqref{eq:fast_subgradient},
$[\mathbf{z}^{(k)}]_j$ depends only on the number of elements in the array
$\{[\mathbf{b}_1]_j, [\mathbf{b}_2]_j, \ldots, [\mathbf{b}_{N}]_j\}$
being less than $[\mathbf{b}^{(k)}]_j$ 
and the number of elements being greater than $[\mathbf{b}^{(k)}]_j$.
Hence, $[\mathbf{z}^{(k)}]_j$ can be computed using a sorting algorithm.
Let $\sigma_j$ be the permutation sorting the array
$\{[\mathbf{b}_1]_j, [\mathbf{b}_2]_j, \ldots, [\mathbf{b}_{N}]_j\}$ in ascending order.
Let $l_j$ be the index at which $[\mathbf{b}^{(k)}]_j$ is inserted into this sorted array while maintaining the ascending order.
We then obtain the following:
\begin{align}
\label{eq:fast_subgradient2}
[\mathbf{z}^{(k)}]_j =-N + 2 l_j - 2.
\end{align}
Appendix \ref{sec:appendix_fast_subgradient} details this derivation.
Note that when there exists an index $i$ such that $[\mathbf{b}^{(k)}]_j=[\mathbf{b}_i]_j$, $l_j$ is not uniquely determined,
but it corresponds to a case in which $\text{sign}(0) \in \{-1, 1\}$,
and $\mathbf{g}^{(k)}$ calculated from $l_j$ is also the subgradient of $f$ at $\mathbf{a}^{(k)}$.
Considering that the permutation $\sigma_j$ does not depend on $[\mathbf{b}^{(k)}]_j$,
$\sigma_j$ can be computed before starting the iterations.
Then, the index $l_j$ is obtained by the binary search, whose time complexity is $O(\log(N))$,
and $\mathbf{z}^{(k)}$ is computed in $O(|\bm{V}| \log(N))$.
Combining $|\bm{V}| < 2|\bm{V}_{\text{leaf}}|$ and the property in which $\mathbf{B}^\top$ is a sparse matrix,
the subgradient $\mathbf{g}^{(k)}$ can be computed in $O(|\bm{V}_{\text{leaf}}|(\log(N) + D))$.

\textbf{Objective function value.}
Next, 
to reduce the time complexity for computing $f(\mathbf{a}^{(k)})$,
we show that a similar way as the above algorithm can be used.
The objective function is rewritten as follows:
\begin{align}
\label{eq:fast_objective_function}
    f(\mathbf{a}^{(k)}) = \frac{1}{N} \sum_{j=1}^{|\bm{V}|} \sum_{i=1}^{N} \left| [\mathbf{b}^{(k)}]_j - [\mathbf{b}_i]_j \right|.
\end{align}
As in the algorithm to compute $\mathbf{g}^{(k)}$,
let $\sigma_j$ be a permutation that sorts the array
$\{[\mathbf{b}_1]_j, [\mathbf{b}_2]_j, \ldots, [\mathbf{b}_{N}]_j\}$
in ascending order.
Let $l_j$ be the index at which $[\mathbf{b}^{(k)}]_j$ is inserted into this sorted array while maintaining the ascending order.
We obtain the following:
\begin{align}
\label{eq:fast_objective_function2}
    &\sum_{i=1}^{N} \left| [\mathbf{b}^{(k)}]_j - [\mathbf{b}_i]_j \right| = \\
    &\left( \sum_{i=1}^{N} [\mathbf{b}_i]_j \right) \!-\! 2 \!\left( \sum_{i=1}^{l_j-1} [\mathbf{b}_{\sigma_j(i)}]_j \right) \!-\! (N - 2l_j + 2) [\mathbf{b}^{(k)}]_j. \nonumber
\end{align}
Note that the second term on the right-hand side is $0$ when $l_j=1$.
The detailed derivation is shown in Appendix \ref{sec:appendix_fast_objective_function2}.
The first term on the right-hand side can be computed before starting the iterations.
The index $l_j$ has already been obtained when computing the subgradient.
Moreover, the second term on the right-hand side can be obtained in $O(1)$ by computing and storing it for all $l_j \in [\![N+1 ]\!]$ before starting the iterations.
Therefore, $f(\mathbf{a}^{(k)})$ can be computed in $O(|\bm{V}|)$.
In summary, using Eqs. \eqref{eq:fast_subgradient2} - \eqref{eq:fast_objective_function2}, the time complexity of the PSD for each iteration can be reduced to $O(|\bm{V}_{\text{leaf}}| (\log(|\bm{V}_{\text{leaf}}|) + \log(N) + D))$,
which is faster than the PSD in terms of the number of samples $N$.
We refer to this algorithm as the \textit{FastPSD}.
Algorithm \ref{alg:fast_psd} shows the FastPSD,
where \textbf{SEARCH} is the function that, given an element and a sorted array, 
returns the index at which the element is inserted into the sorted array while maintaining the ascending order.

\begin{algorithm}[!t]
   \caption{FastPSD for the FS-TWB.}
   \label{alg:fast_psd}
\begin{algorithmic}[1]
   \STATE {\bfseries Input:} Probability measures $\mathbf{a}_1, \mathbf{a}_2, \ldots, \mathbf{a}_N$, and step size $0 < \gamma_1$ and $0< \gamma_2 \leq 1$.
   \STATE {\bfseries Output:} The FS-TWB.
   \FOR{$i = 1, 2, \ldots, N$}
   \STATE $\mathbf{b}_i \leftarrow \mathbf{B} \mathbf{a}_i$
   \ENDFOR
   \FOR{$j = 1, 2, \ldots, |\bm{V}|$}
   \STATE Compute and store the permutation $\sigma_j$ that sorts the array $\{[\mathbf{b}_i]_j\}_{i=1}^{N}$.
   \FOR{$l = 1, 2, \ldots, N+1$}
   \STATE Compute and store $\sum_{i=1}^{l-1} [\mathbf{b}_{\sigma_j(i)}]_j$.
   \ENDFOR
   \STATE Compute and store $\sum_{i=1}^{N} [\mathbf{b}_i]_j$.
   \ENDFOR
   \STATE Let $\mathbf{a}^{(0)} \in \bm{A}$. 
   \STATE $\mathbf{a}^{\text{best}} \leftarrow \mathbf{a}^{(0)}$ 
   \STATE $\mathbf{b}^{(0)} \leftarrow \mathbf{B} \mathbf{a}^{(0)}$
   \STATE Compute $f^{(0)}$ by Eqs. \eqref{eq:fast_objective_function} and \eqref{eq:fast_objective_function2}.
   \STATE $f^{\text{best}} \leftarrow f^{(0)}$
   \FOR{$k = 0, 1, \ldots, K$}
   \FOR{$j = 1, 2, \ldots, |\bm{V}|$}
   \STATE $l_j \leftarrow \textbf{SEARCH}([\mathbf{b}^{(k)}]_j, \{ [\mathbf{b}_{\sigma_j(i)}]_j \}_{i=1}^{N})$
   \ENDFOR
   \STATE Compute $\mathbf{z}^{(k)}$ by Eq. \eqref{eq:fast_subgradient2}.
   \STATE $\mathbf{g}^{(k)} \leftarrow \frac{1}{N} \mathbf{B}^\top \mathbf{z}^{(k)}$
   \STATE $\gamma^{(k)} \leftarrow \frac{\gamma_1}{(k + 1)^{\gamma_2} \| \mathbf{g}^{(k)} \|_2}$
   \STATE $\mathbf{a}^{(k + 1)} \leftarrow \text{proj}_{\bm{A}}(\mathbf{a}^{(k)} - \gamma^{(k)} \mathbf{g}^{(k)})$
   \STATE $\mathbf{b}^{(k+1)} \leftarrow \mathbf{B} \mathbf{a}^{(k+1)}$
   \STATE Compute $f^{(k+1)}$ by Eqs. \eqref{eq:fast_objective_function} and \eqref{eq:fast_objective_function2}. 
   \IF {$f^{\text{best}} > f^{(k+1)}$}
   \STATE $\mathbf{a}^{\text{best}} \leftarrow \mathbf{a}^{(k+1)}$
   \STATE $f^{\text{best}} \leftarrow f^{(k+1)}$
   \ENDIF
   \ENDFOR
   \STATE {\bfseries return} $\mathbf{a}^{\text{best}}$
\end{algorithmic}
\end{algorithm}

\subsection{Fixed Support Tree-Sliced Wasserstein Barycenter}
\label{sec:fs-tswb}
In this section, we propose an extension of the FS-TWB,
the barycenter under the tree-sliced Wasserstein distance \citep{le2019tree},
and show that the PSD and the FastPSD can be naturally applied to solve it.

Let $T$ be the number of sampled tree metrics, and let $\{ d_{\mathcal{T}^{(t)}}\}_{t=1}^{T}$ be a set of sampled tree metrics.
The tree-sliced Wasserstein distance is defined as follows:
\begin{align}
    W_{\overline{d}_{\mathcal{T}}} (\mu_i, \mu_j) = \frac{1}{T} \sum_{t=1}^{T} W_{d_{\mathcal{T}^{(t)}}} (\mu_i, \mu_j).
\end{align}
\citet{le2019tree} showed that the tree-sliced Wasserstein distance can better approximate the Wasserstein distance when the number of trees increases.
In the previous sections, we discuss the case in which $T=1$.
Then, the barycenter under the tree-sliced Wasserstein distance is defined as follows:
\begin{align}
\label{eq:fs-tswb}
    \overline{\mu}_{\overline{d}_{\mathcal{T}}} \in \argmin_{\mu \in P(\bm{V}_{\text{leaf}})} \frac{1}{N} \left( \sum_{i=1}^{N} W_{\overline{d}_{\mathcal{T}}} (\mu, \mu_i) \right),
\end{align}
which we refer to as the \textit{fixed support tree-sliced Wasserstein barycenter} (FS-TSWB).
Because this objective function is the average of the objective functions of the FS-TWB problem,
it is a nondifferential convex function and Lipschitz continuous.
Therefore, the FS-TSWB problem is also a convex nondifferentiable optimization,
which can be solved using the PSD and the FastPSD.
More specifically, the subgradient of the objective function of the FS-TSWB problem can be obtained 
as the average of the subgradients of the objective function of the FS-TWB problem.
Then, the subgradient and the objective function value of the FS-TSWB problem can be computed fast as in Algorithm \ref{alg:fast_psd},
whose time complexity for each iteration is $O(T |\bm{V}_{\text{leaf}}| (\log(|\bm{V}_{\text{leaf}}|) + \log(N) + D))$.
Moreover, because a chain is a tree, the PSD and the FastPSD can solve the fixed support sliced Wasserstein barycenter (FS-SWB) problem.
\citet{julien2011wasserstein} and \citet{bonneel2015sliced} have studied the sliced Wasserstein barycenter only in the free support setting.
To the best of our knowledge, our study is the first to propose an algorithm for solving the FS-SWB problem.
Appendix \ref{sec:fastpsd_fot_fs-swb} details the method for applying the FastPSD to the FS-SWB problem.

\section{Experiment}
In this section, 
we evaluate the FS-TSWB and the FastPSD on MNIST, AMAZON, and AGNews.

\subsection{Datasets}
MNIST contains $60000$ handwritten digit images, which are categorized into ten groups.
Similar to the previous work \citep{cuturi2014fast},
images are considered as the distributions on $28\times28$ pixels.
We use the two-dimensional Euclidean distances between each pixel location as the ground metric.
AMAZON consists of approximately $8000$ documents pre-processed by the previous works \citep{kusner2015from}.
The documents are categorized into four groups,
and each category contains approximately $13000$ unique words on average.
AGNews consists of approximately $120000$ documents, which are categorized into four groups.
We remove the stop words and stem the words.
Each category then contains approximately $13000$ unique words on average.
On AMAZON and AGNews, we use GloVe \citep{pennington2014global}, which is $50$ dimensions and pre-trained on Wikipedia, as the ground metric.

\subsection{Comparison Methods}
\textbf{Fixed Support Wasserstein Barycenter (FS-WB):}
To solve the FS-WB problem in Eq. \eqref{eq:fs_wb}, we use the IBP \citep{benamou2015iterative} as the baseline method
\footnote{We evaluated \citep{dvinskikh2021improved} as an additional baseline to solve the FS-WB by using the implementation contained in their supplementary material. However, in practice, the IBP is faster. Therefore, we only show the results of the IBP.}.
We set the entropic regularization parameter to $0.01$, 
the maximum iteration to $1000$, and the threshold for the stopping criteria to $0.0001$.
We use the public implementation\footnote{\url{https://pythonot.github.io/}},
which is written with Python.

\textbf{Fixed Support Tree-Sliced Wasserstein Barycenter (FS-TSWB):}
To sample the trees, we use the farthest point clustering method \citep{le2019tree},
and for all $v \in \bm{V}\setminus \{ v_1 \}$, we set edge length $w_v$ to one.
The depth of the tree is set to $6$, and the number of child nodes is set to $5$.
For the fast convergence, 
we set the initial value to $\mathbf{a}^{(0)} = \frac{1}{N} \sum_{i=1}^N \mathbf{a}_i$.
We set the step size $\gamma_1 = 0.05$ and $\gamma_2 = 0.25$, and the iteration number to $1500$.
The number of sampled trees $T$ is set to $1, 5, 10, 15, 20$, and $25$.
We implement the PSD and the FastPSD using Python.

\textbf{Fixed Support Sliced Wasserstein Barycenter (FS-SWB):}
We use the FastPSD to solve the FS-SWB problem
and set the parameters of the FastPSD to the same values as those of the FS-TSWB.

When evaluating the time consumption to compute the barycenters,
we run all methods on Intel Xeon Gold 6226R CPU @ 2.90GHz
where the maximum number of threads is limited to eight.

\subsection{Numerical Results}
\begin{figure*}[t!]
\vskip -0.15 in
\subfigure[MNIST]{
\includegraphics[width=0.3\textwidth]{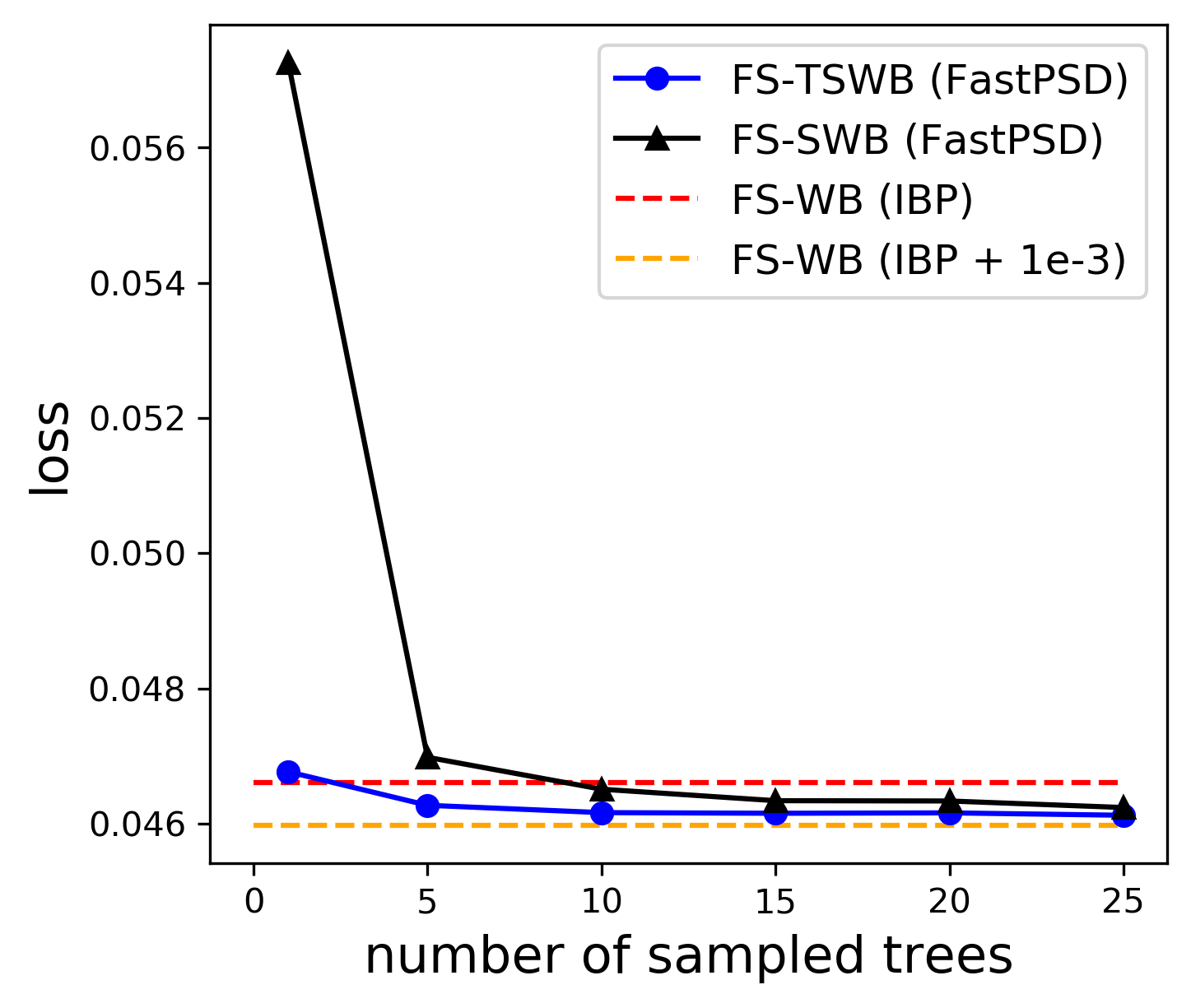}
}
\subfigure[AMAZON]{
\includegraphics[width=0.3\textwidth]{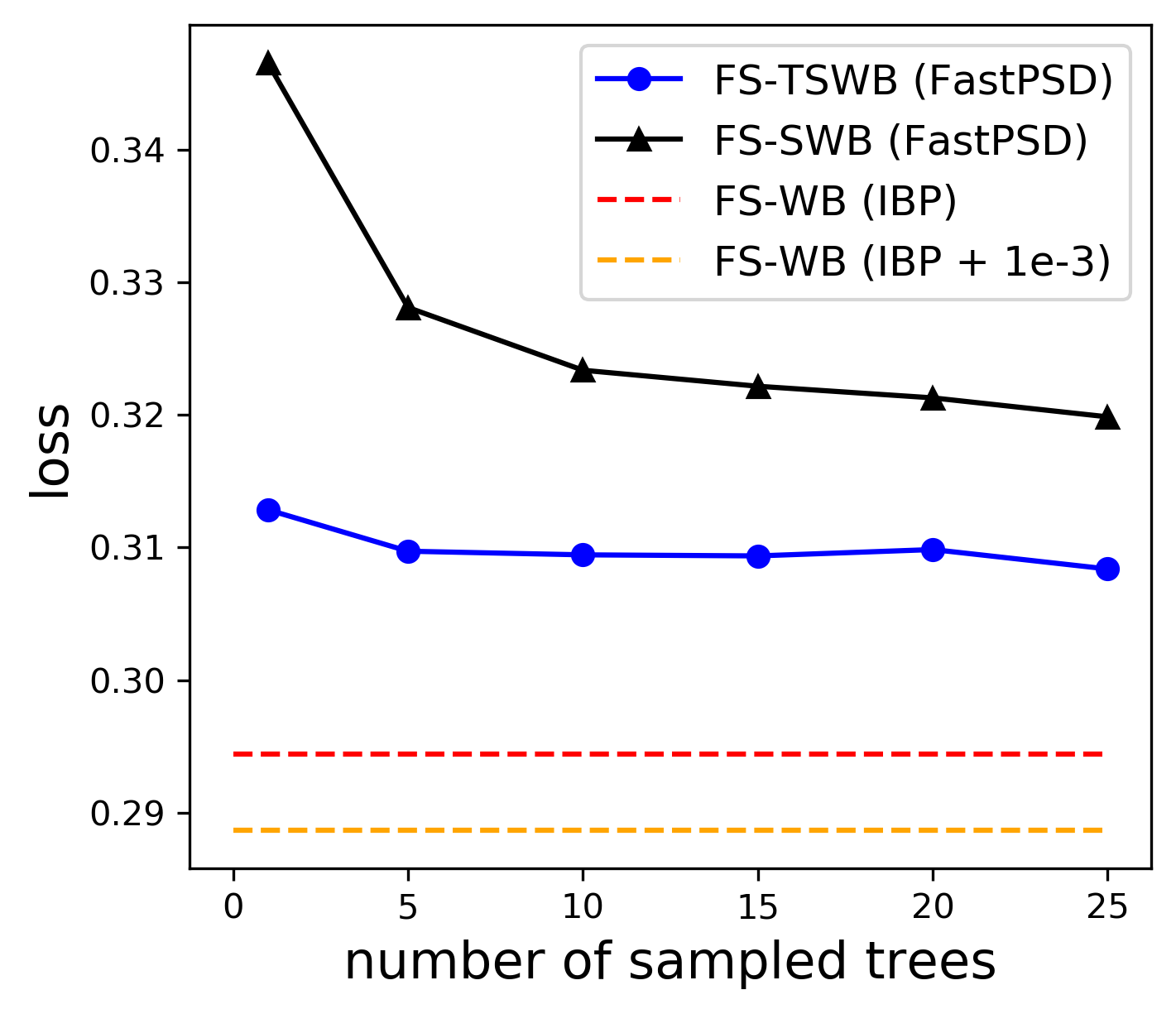}
}
\subfigure[AGNews]{
\includegraphics[width=0.3\textwidth]{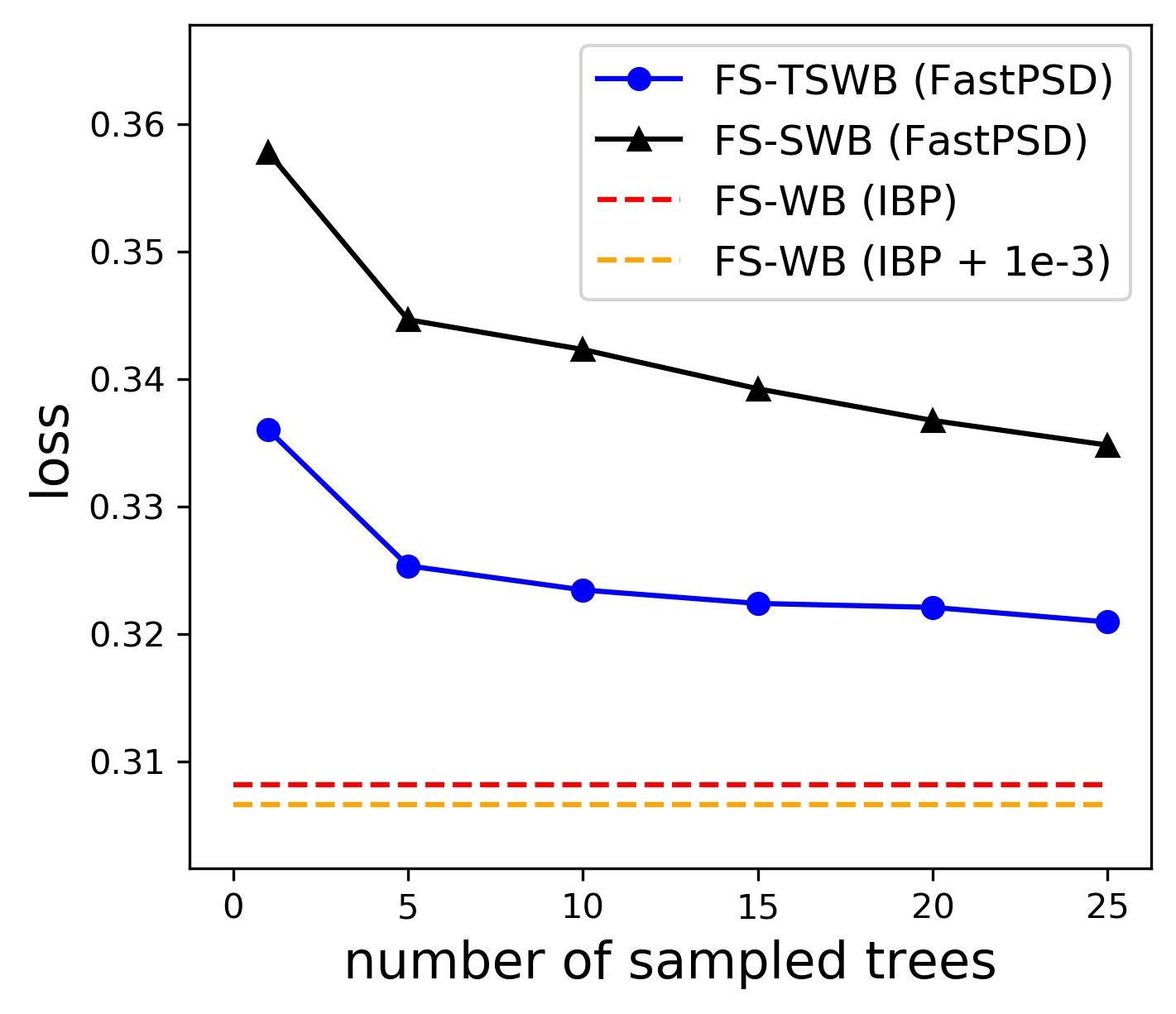}
}
\vskip -0.15 in
\caption{Objective function value for Eq. \eqref{eq:fs_wb} with the entropic regularization.
The results are averages for all categories.
To compute the loss at each barycenter, we use the Sinkhorn algorithm with the same parameters as the IBP.
The FS-WB (IBP + 1e-3) denotes the barycenter normalized such that the sum is one after all probabilities in the FS-WB (IBP) less than $0.001$ are set to zero.}
\vskip -0.1 in
\label{fig:wb_loss}
\end{figure*}
In this section, we evaluate the FS-TSWB using the objective function value of the FS-WB.
In the following, we refer to \textit{loss} as the objective function value of the FS-WB problem.
For example, the loss at the FS-TSWB denotes $\frac{1}{N} \sum_i W_d(\mu_i, \overline{\mu}_{\overline{d}_{\mathcal{T}}})$.
Fig. \ref{fig:wb_loss} shows the loss at the FS-WB, the FS-SWB, and the FS-TSWB.
Comparing the FS-SWB and the FS-TSWB, the loss at the FS-TSWB is smaller than the loss at the FS-SWB on all datasets.
The reason is that, because a tree has more degrees of freedom than a chain, 
a tree can approximate the original space better than a chain.
Next, we compare the FS-WB and the FS-TSWB.
On all datasets, the loss at the FS-TSWB decreases as the number of sampled trees increases.
In particular, on MNIST, as the number of sampled trees increases,
the loss at the FS-TSWB becomes smaller than the loss at the FS-WB obtained by the IBP.
Because there are many pixels on which the probability is zero in all images,
the probability on many pixels is zero in the optimal FS-WB.
However, in practice, the probability on these pixels are not zero in the FS-WB obtained by the IBP.
Indeed, the result shows that, in the FS-WB obtained by the IBP, 
the loss decreases by setting the probability below the threshold to zero.
On the other hand, in the FS-TSWB obtained by the FastPSD, 
the probability on these pixels is zero by the projection onto the simplex per iteration. 
As a result, the loss at the FS-TSWB obtained by the FastPSD is smaller than the loss at the FS-WB obtained by the IBP.

\subsection{Visualization of Barycenters}
\begin{figure}[t]
\vskip -0.0 in
\begin{center}
\centerline{\includegraphics[width=\columnwidth]{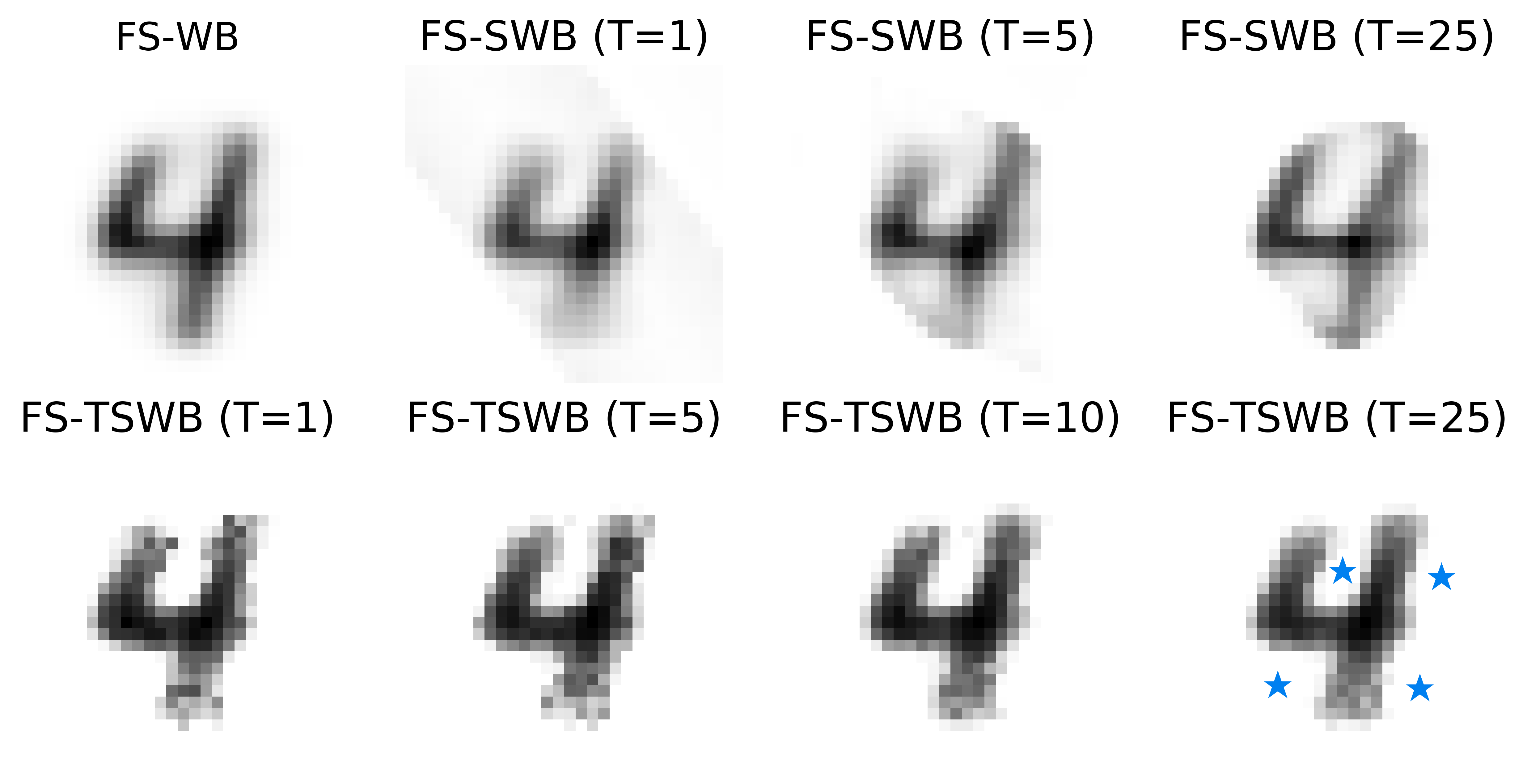}}
\vskip -0.1 in
\caption{Visualization of the FS-WB, the FS-SWB, and the FS-TSWB on MNIST.}
\label{fig:mnist}
\end{center}
\vskip -0.4 in
\end{figure}

In this section, we show a visualization of the barycenters.
Fig. \ref{fig:mnist} shows the FS-WB, the FS-SWB, and the FS-TSWB on MNIST.
Comparing the FS-SWB and the FS-TSWB, the FS-TSWB is closer to the FS-WB than the FS-SWB.
In the FS-SWB, some pixels have an unnaturally high probability.
In particular, the pixels in the area indicated by the blue stars in Fig. \ref{fig:mnist} 
have a high probability even if the number of chains increases.
By contrast,
in the FS-TSWB, the probability on the pixels in the area indicated by the blue stars is properly zero
even if the number of trees is one.
Moreover, the results show that increasing the number of the trees can make the FS-TSWB smoother.
Appendix \ref{sec:other_visualization} includes the remaining visualization of the barycenters.

\subsection{Time Consumption}
In this section, 
we evaluate the time consumption of the FS-TSWB.
Table \ref{table:time} shows the time required to solve the FS-WB and FS-TSWB problems by using the IBP and the FastPSD respectively.
When the number of sampled trees is one,
the FS-TSWB can be solved faster than the FS-WB on all datasets.
In particular, on AGNews, using the FastPSD, the FS-TSWB can be solved approximately $125$ times faster than the FS-WB.
Comparing the time consumption of the FS-TSWB when the number of trees increases,
the time consumption of the FS-TSWB increases in proportion to the number of sampled trees.
Then, there is the trade off between the performance and the time consumption.

\begin{figure}[t]
\subfigure{
\includegraphics[width=0.45\hsize]{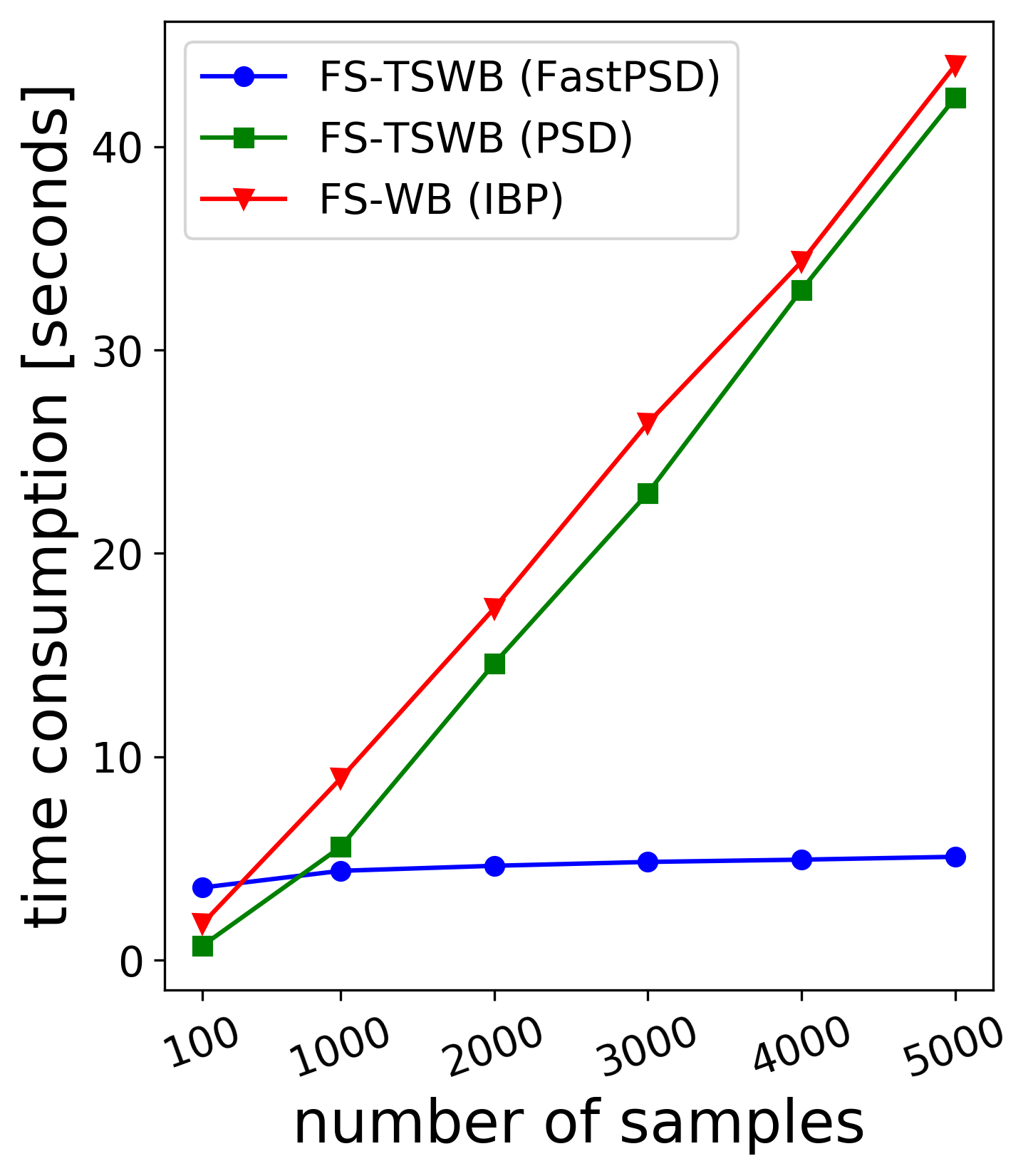}
}
\subfigure{
\includegraphics[width=0.46\hsize]{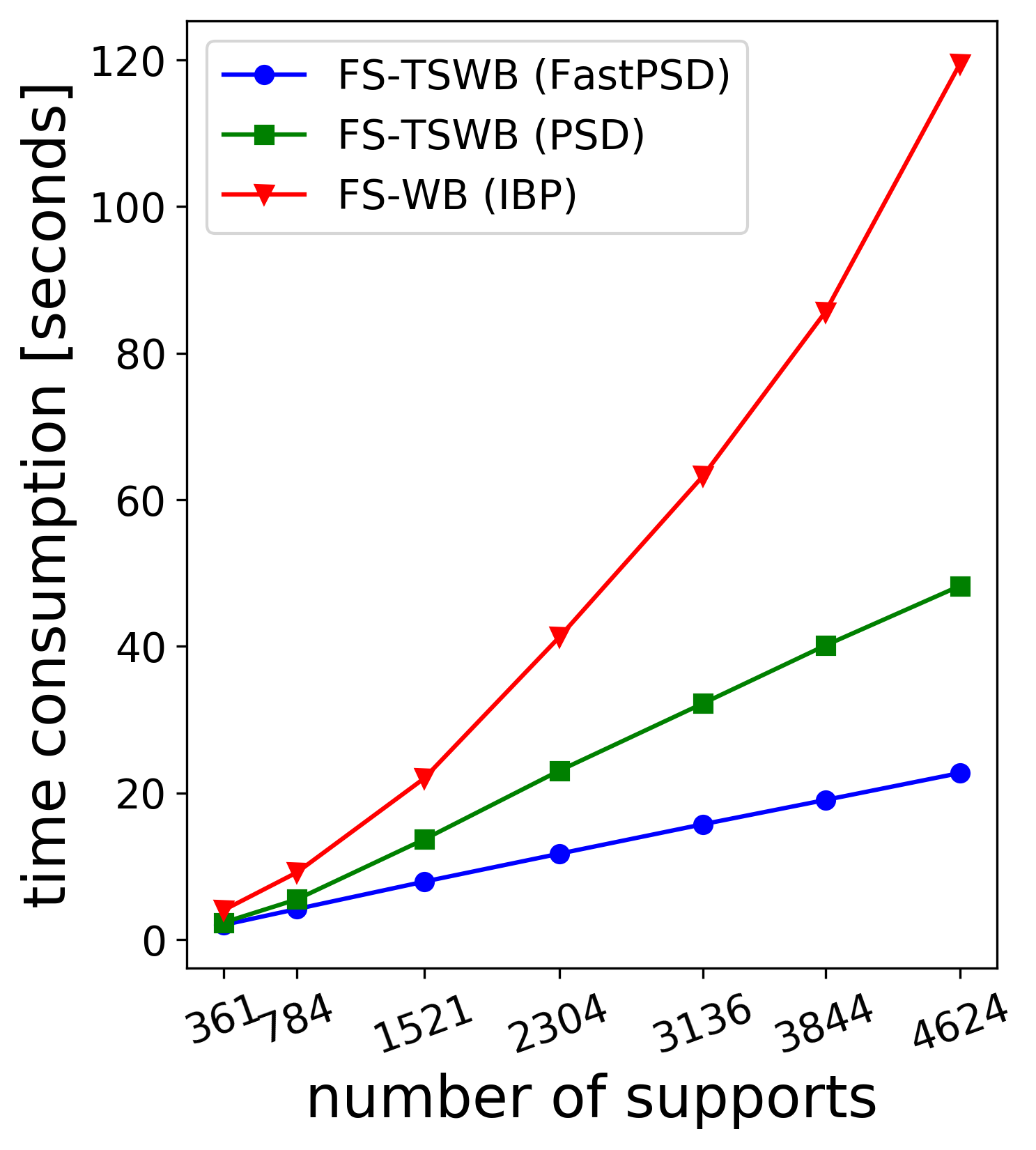}
}
\vskip -0.15 in
\caption{Time consumption when varying the number of samples and the number of supports on MNIST. The number of tree is set to one. When the number of supports is varied, the number of samples is set to $1000$. The results are averages for all categories.}
\vskip -0.15 in
\label{fig:speed}
\end{figure}

In addition, 
we evaluate the time consumption in more details on MNIST.
Fig. \ref{fig:speed} shows the time consumption when varying the number of images and
when varying the number of supports by resizing the image.
The results show that the time consumption of the IBP and the PSD increases linearly with respect to the number of samples.
By contrast, the time consumption of the FastPSD increases with $O(\log(N))$.
As a result, the time consumption of the FastPSD is almost the same even if the number of samples increases.
Next, we compare the results when varying the number of supports.
The results show that the time consumption of the IBP increases quadratically with respect to the number of supports.
By contrast, the time consumption of the PSD and the FastPSD increases with $O(|\bm{V}_{\text{leaf}}| \log (|\bm{V}_{\text{leaf}}|))$.
In summary, using the FastPSD, the FS-TSWB problem can be solved faster than the FS-WB problem 
in terms of both the number of samples and the number of supports.
\begin{table}[t!]
\vskip -0.25 in
\caption{Time consumption [seconds].} 
\vskip -0.1 in
\label{table:time}
\small
\begin{center}
\begin{tabular}{cccc}
& MNIST & AMAZON & AGNews \\ \hline
FS-WB            &  64.4 & 2129.6 & 10811.7 \\
FS-TSWB ($T=1$)  &   5.2 &   62.4 &    86.1 \\ 
FS-TSWB ($T=5$)  &  25.1 &  330.7 &   449.4 \\
FS-TSWB ($T=10$) &  51.1 &  653.9 &   899.9 \\ 
FS-TSWB ($T=15$) &  78.9 &  969.5 &  1346.6 \\
FS-TSWB ($T=20$) & 111.9 & 1287.3  & 1788.4 \\
FS-TSWB ($T=25$) & 142.6 & 1610.3  & 2236.4 \\ \hline
\end{tabular}
\end{center}
\vskip -0.1 in
\end{table}

\subsection{Memory Consumption}
\begin{table}[t]
\vskip -0.25 in
\caption{Peak memory consumption [GB].} 
\vskip -0.1 in
\label{table:memory}
\small
\begin{center}
\begin{tabular}{cccc}
& MNIST & AMAZON & AGNews \\ \hline
FS-WB            & 0.41 & 4.30 &  16.22 \\
FS-TSWB ($T=1$)  & 0.39 & 0.99 &  12.25 \\ 
FS-TSWB ($T=5$)  & 0.71 & 2.28 &  30.57 \\
FS-TSWB ($T=10$) & 1.10 & 3.91 &  53.50 \\ 
FS-TSWB ($T=15$) & 1.50 & 5.53 &  76.30 \\
FS-TSWB ($T=20$) & 1.87 & 7.15 &  99.14 \\
FS-TSWB ($T=25$) & 2.25 & 8.75 & 121.91 \\ \hline
\end{tabular}
\end{center}
\vskip -0.1 in
\end{table}

\begin{figure}[t]
\subfigure{
\includegraphics[width=0.46\hsize]{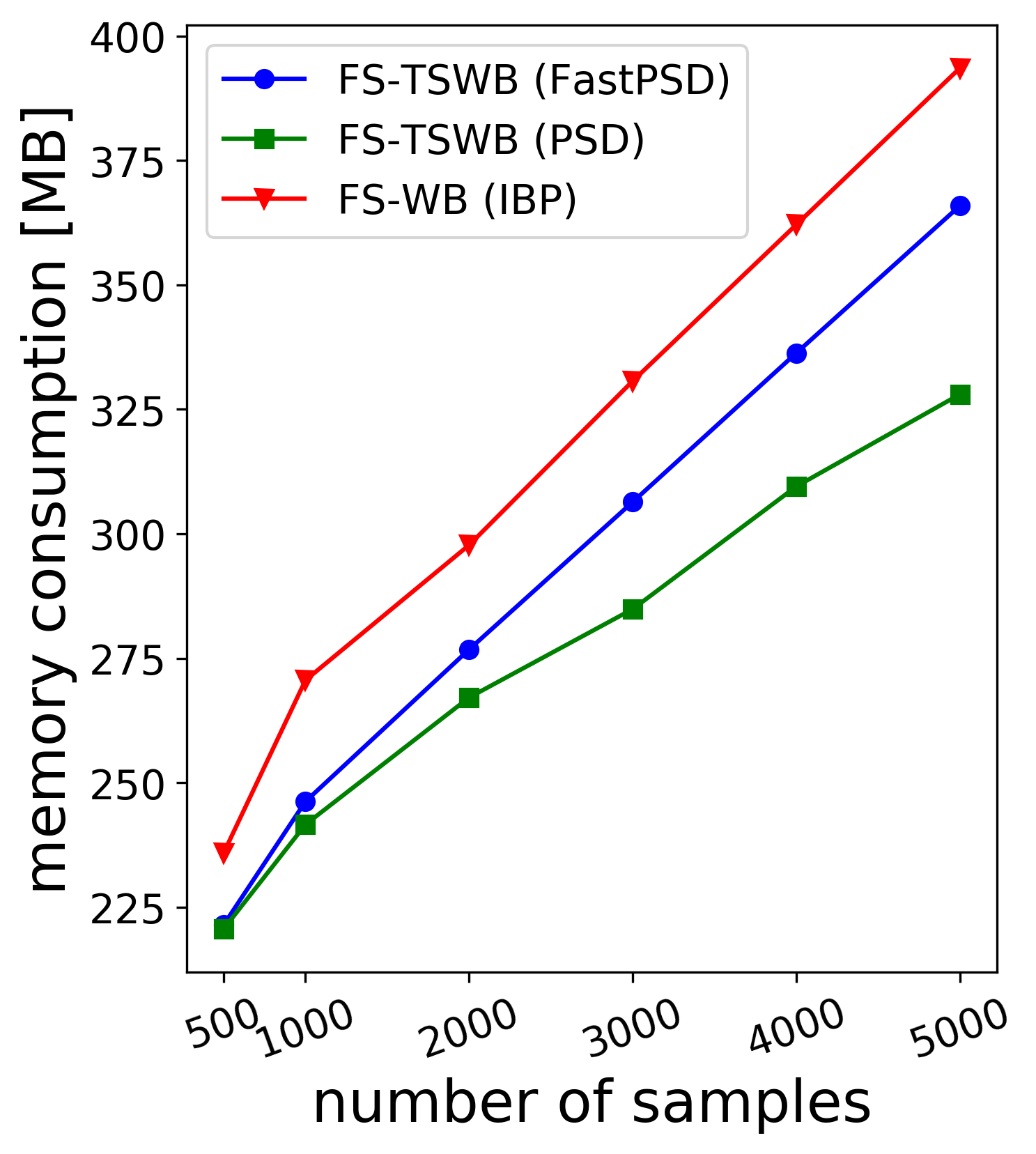}
}
\subfigure{
\includegraphics[width=0.46\hsize]{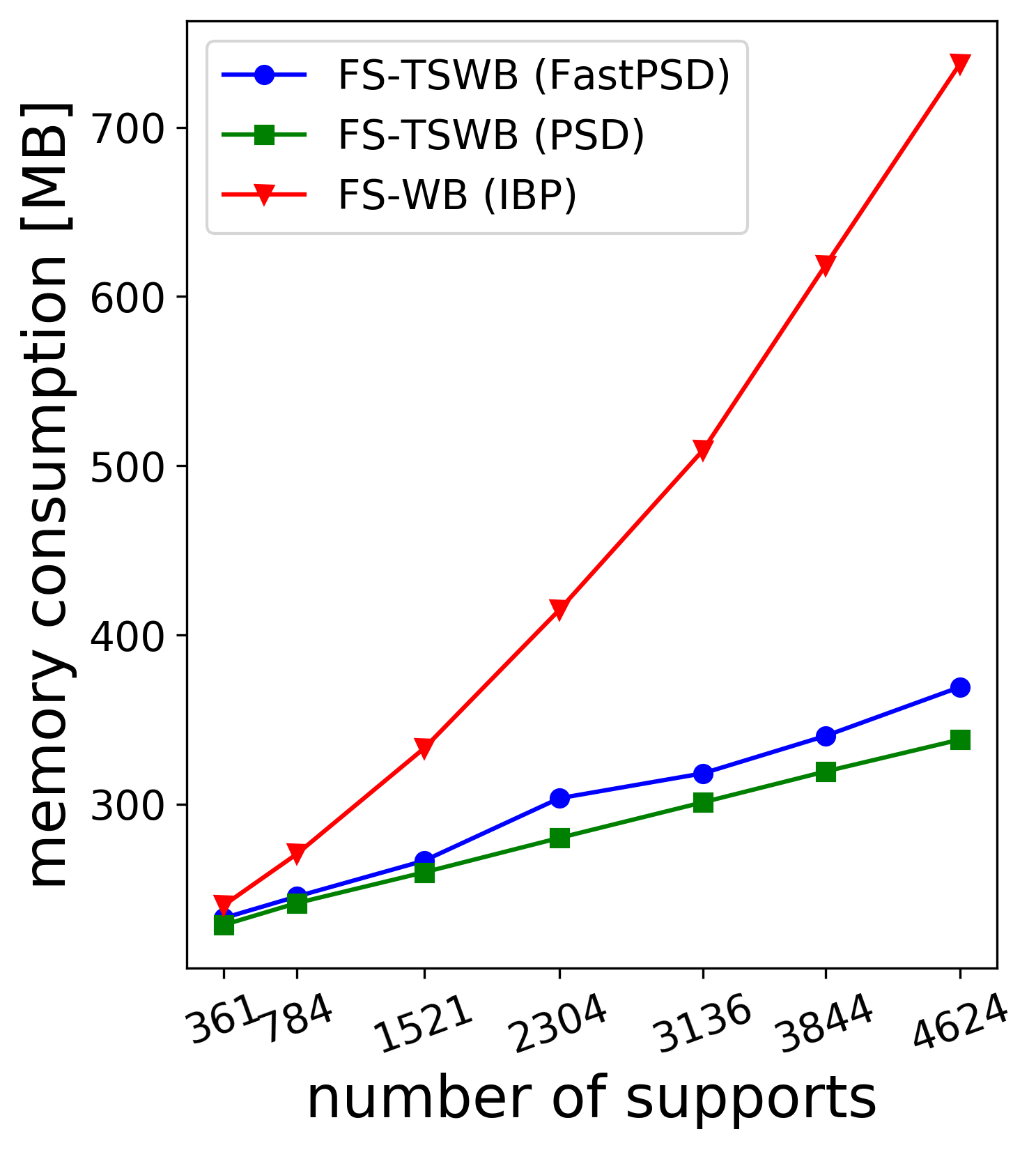}
}
\vskip -0.15 in
\caption{Peak memory consumption when varying the number of samples and the number of supports on MNIST. 
The number of tree is set to one. When the number of supports is varied, the number of samples is set to $1000$.
The results are averages for all categories.}
\vskip -0.15 in
\label{fig:memory}
\end{figure}

In this section, following the previous work \citep{le2020treewasserstein}, 
we evaluate the memory consumption of the FastPSD. 
Table \ref{table:memory} shows the peak memory consumption required to 
solve the FS-WB and FS-TSWB problems by using the IBP and the FastPSD respectively.
The results show that, on all datasets, when the number of sampled trees is one,
the FS-TSWB problem can be solved with less memory consumption than the FS-WB problem.
However, the memory consumption of the FS-TSWB problem increases linearly when the number of sampled trees increases.
The reason is that, for a fast computation, we compute and store $\mathbf{B} \mathbf{a}_i$ for all sampled trees before starting the iterations.

In addition, we evaluate the memory consumption in more details on MNIST.
Fig. \ref{fig:memory} shows the memory consumption when varying the number of samples and
when varying the number of supports.
When the number of samples increases, the memory consumption of all methods increases linearly.
When the number of supports increases, althouth the memory consumption of the IBP increases quadratically,
the memory consumption of the PSD and the FastPSD increase linearly.
The reason is that the IBP uses the $|\bm{V}_{\text{leaf}}| \times |\bm{V}_{\text{leaf}}|$ cost matrix,
while the PSD and the FastPSD use the sparse matrix $\mathbf{B}$ instead of this cost matrix.
As a result, when the number of supports is large, the FS-TSWB problem can be solved with less memory consumption than the FS-WB problem.

\section{Conclusion}
In this paper,
we properly formulate the barycenter under the tree-Wasserstein distance, called the FS-TWB, and its extension, called the FS-TSWB.
We then propose an efficient optimization algorithm to solve these problems.
Specifically, 
we show that the FS-TWB and FS-TSWB problems are convex optimizations, which can be solved using the PSD.
Moreover, by using the properties of these problems, 
we propose a more efficient algorithm to compute the subgradient and the objective function value, called the FastPSD.
Experimental results show that, by using the FastPSD, 
the FS-TWB and FS-TSWB problems can be solved extremely faster than the FS-WB problem with less memory consumption.
In addition, comparing the FS-SWB and the FS-TSWB, 
we show that the FS-TSWB can approximate the FS-WB better than the FS-SWB.
Furthermore, the results show that by sampling multiple trees, the FS-TSWB becomes closer to the FS-WB.

\section*{Acknowledgments}
M.Y. was supported by MEXT KAKENHI Grant Number 20H04243. 
R.S. was supported by JSPS KAKENHI Grant Number 21J22490.

\bibliography{example_paper}
\clearpage
\appendix
\thispagestyle{empty}

\onecolumn\makesupplementtitle

\section{Summary of Time Complexity}
\begin{table}[H]
\begin{center}
\vskip -0.2 in
\caption{Time complexity per iteration.}
\vskip 0.1 in
\label{table:time_complexity}
\begin{tabular}{lc}
\toprule
                  & Time Complexity \\ 
\midrule
FS-WB (IBP)       & $O(N |\mathbf{V}_{\text{leaf}}|^2)$ \\ 
FS-TSWB (PSD)     & $O(T|\mathbf{V}_{\text{leaf}}| ( \log (|\mathbf{V}_{\text{leaf}}|) + N + D))$ \\ 
FS-TSWB (FastPSD) & $O(T|\mathbf{V}_{\text{leaf}}| ( \log (|\mathbf{V}_{\text{leaf}}|) + \log (N) + D))$ \\ 
\bottomrule
\end{tabular}
\end{center}
\end{table}

\section{Derivation of Eq. \eqref{eq:fast_subgradient2}}
\label{sec:appendix_fast_subgradient}
We obtain the following:
\begin{align*}
    [\mathbf{z}^{(k)}]_j &= \sum_{i=1}^{N} \text{sign} \left( [\mathbf{b}^{(k)}]_j - [\mathbf{b}_i]_j \right) \\
    &= \sum_{i=1}^{N} \text{sign} \left( [\mathbf{b}^{(k)}]_j - [\mathbf{b}_{\sigma_j(i)}]_j \right).
\end{align*}
Then,
if $l_j=1$, we obtain the following:
\begin{align*}
    \sum_{i=1}^{N} \text{sign} \left( [\mathbf{b}^{(k)}]_j - [\mathbf{b}_{\sigma_j(i)}]_j \right) &= \sum_{i=1}^{N} -1 = -N.
\end{align*}
If $l_j=N+1$, we obtain the following:
\begin{align*}
    \sum_{i=1}^{N} \text{sign} \left( [\mathbf{b}^{(k)}]_j - [\mathbf{b}_{\sigma_j(i)}]_j \right) &= \sum_{i=1}^{N} 1 = N.
\end{align*}
If $2\leq l_j \leq N$, we obtain the following:
\begin{align*}
    \sum_{i=1}^{N} \text{sign} \left( [\mathbf{b}^{(k)}]_j - [\mathbf{b}_{\sigma_j(i)}]_j \right) &= \sum_{i=1}^{l_j - 1} 1 -  \sum_{i=l_j}^{N} 1 \\
    &= l_j - 1 - (N - l_j + 1) \\ 
    &= - N + 2 l_j - 2.
\end{align*}

Therefore, for any $l_j \in [\![ N+1 ]\!]$, we obtain the following:
\begin{align*}
    [\mathbf{z}^{(k)}]_j = - N + 2 l_j - 2.
\end{align*}

\section{Derivation of Eq. \eqref{eq:fast_objective_function2}}
\label{sec:appendix_fast_objective_function2}
We obtain the following:
\begin{align*}
    \sum_{i=1}^{N} \left| [\mathbf{b}^{(k)}]_j - [\mathbf{b}_i]_j \right| &= \sum_{i=1}^{N} \left| [\mathbf{b}^{(k)}]_j - [\mathbf{b}_{\sigma_j(i)}]_j\right|.
\end{align*}
Then, if $l_j=1$, we obtain
\begin{align*}
    \sum_{i=1}^{N} \left| [\mathbf{b}^{(k)}]_j - [\mathbf{b}_{\sigma_j(i)}]_j\right| &= \sum_{i=1}^{N} -[\mathbf{b}^{(k)}]_j + [\mathbf{b}_{\sigma_j(i)}]_j \\
    &= - \left( \sum_{i=1}^{N} [\mathbf{b}^{(k)}]_j \right) + \left( \sum_{i=1}^{N} [\mathbf{b}_{\sigma_j(i)}]_j \right) \\
    &= - N [\mathbf{b}^{(k)}]_j + \left( \sum_{i=1}^{N} [\mathbf{b}_i]_j \right).
\end{align*}
If $l_j=N+1$, we obtain the following:
\begin{align*}
    \sum_{i=1}^{N} \left| [\mathbf{b}^{(k)}]_j - [\mathbf{b}_{\sigma_j(i)}]_j\right| &= \sum_{i=1}^{N} [\mathbf{b}^{(k)}]_j - [\mathbf{b}_{\sigma_j(i)}]_j \\
    &= \left( \sum_{i=1}^{N} [\mathbf{b}^{(k)}]_j \right) - \left( \sum_{i=1}^{N} [\mathbf{b}_{\sigma_j(i)}]_j \right) \\
    &= N [\mathbf{b}^{(k)}]_j - \left( \sum_{i=1}^{N} [\mathbf{b}_i]_j \right).
\end{align*}
If $2\leq l_j \leq N$, we obtain
\begin{align*}
    \sum_{i=1}^{N} \left| [\mathbf{b}^{(k)}]_j - [\mathbf{b}_{\sigma_j(i)}]_j\right| &= \sum_{i=1}^{l_j-1} \left( [\mathbf{b}^{(k)}]_j - [\mathbf{b}_{\sigma_j(i)}]_j \right) - \sum_{i=l_j}^{N} \left( [\mathbf{b}^{(k)}]_j - [\mathbf{b}_{\sigma_j(i)}]_j \right) \\
    &= \left( l_j - 1 \right) [\mathbf{b}^{(k)}]_j  - \left(\sum_{i=1}^{l_j-1} [\mathbf{b}_{\sigma_j(i)}]_j \right)- \left(N - l_j + 1 \right) [\mathbf{b}^{(k)}]_j + \left( \sum_{i=l_j}^{N} [\mathbf{b}_{\sigma_j(i)}]_j \right) \\
    &= \left( \sum_{i=l_j}^{N} [\mathbf{b}_{\sigma_j(i)}]_j \right) - \left(\sum_{i=1}^{l_j-1} [\mathbf{b}_{\sigma_j(i)}]_j \right)- \left(N - 2l_j + 2 \right) [\mathbf{b}^{(k)}]_j  \\
    &= \left( \sum_{i=1}^{N} [\mathbf{b}_{\sigma_j(i)}]_j \right) - 2\left(\sum_{i=1}^{l_j-1} [\mathbf{b}_{\sigma_j(i)}]_j \right)- \left(N - 2l_j + 2 \right) [\mathbf{b}^{(k)}]_j \\
    &= \left( \sum_{i=1}^{N} [\mathbf{b}_i]_j \right) - 2\left(\sum_{i=1}^{l_j-1} [\mathbf{b}_{\sigma_j(i)}]_j \right)- \left(N - 2l_j + 2 \right) [\mathbf{b}^{(k)}]_j. \\
\end{align*}
Therefore, for any $l_j \in [\![ N+1 ]\!]$, we obtain the following:
\begin{align*}
    \sum_{i=1}^{N} \left| [\mathbf{b}^{(k)}]_j - [\mathbf{b}_i]_j \right| = \left( \sum_{i=1}^{N} [\mathbf{b}_i]_j \right) - 2\left(\sum_{i=1}^{l_j-1} [\mathbf{b}_{\sigma_j(i)}]_j \right)- \left(N - 2l_j + 2 \right) [\mathbf{b}^{(k)}]_j.
\end{align*}

\section{Fixed Support Sliced Wasserstein Barycenter}
\label{sec:fastpsd_fot_fs-swb}
Because a chain is a tree, the FastPSD can solve the fixed support sliced Wasserstein barycenter (FS-SWB) problem.
However, since the depth of the chain $D$ is $O(|\bm{V}_{\text{leaf}}|)$, the time consumption of the FastPSD increases with $O(|\bm{V}_{\text{leaf}}|^2)$.
In this section, we propose a method for reducing the time complexity to $O(T |\bm{V}_{\text{leaf}}| (\log(|\bm{V}_{\text{leaf}}|) + \log(N)))$.

\subsection{Problem Setting}
\begin{figure}[H]
  \begin{minipage}[b]{.5\columnwidth}
    \raggedleft
    \includegraphics[width=.5\hsize]{{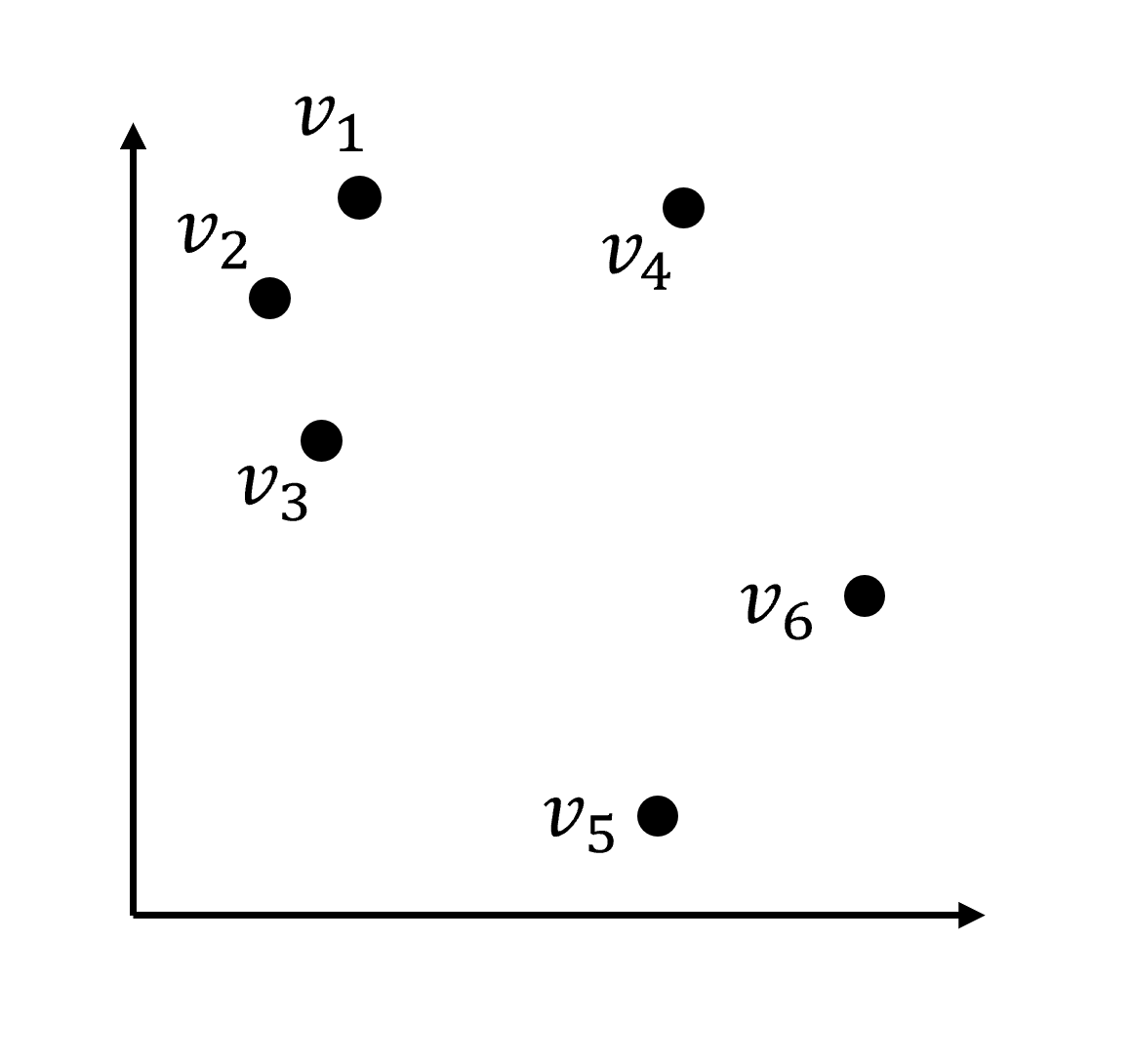}}%
  \end{minipage}%
  \begin{minipage}[b]{.5\columnwidth}
    \raggedright
    \includegraphics[width=.5\hsize]{{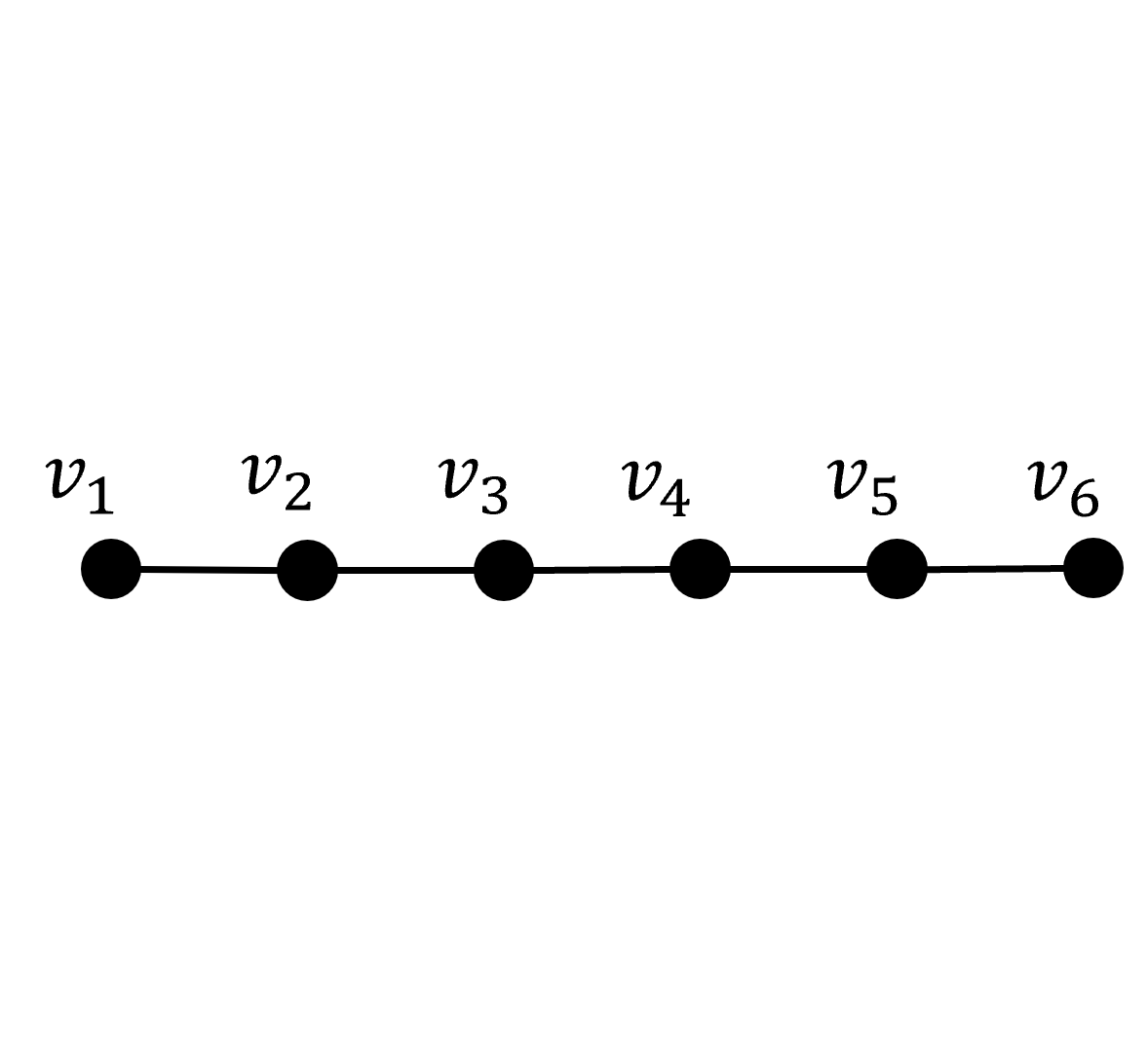}}
  \end{minipage}
\caption{Illustration of the original space (left) and the chain (right).}
\label{fig:illustration_chain}
\end{figure}
In this section, we show the FS-TWB problem when the tree $\mathcal{T}$ is a chain.
In the chain, because nodes that have no corresponding elements in the original space $\Omega$ can be abbreviated, 
all nodes have corresponding elements in $\Omega$.
(i.e., $\bm{V} = \bm{V}_{\text{leaf}}=\Omega$ and $\bm{V}_{\text{in}}=\emptyset$.)
Fig. \ref{fig:illustration_chain} shows an illustration of the chain.
Therefore, when the tree $\mathcal{T}$ is a chain, the FS-TWB problem is equivalent to the following:
\begin{align}
    \overline{\mu} \in \argmin_{\mu \in P(\bm{V})} \frac{1}{N} \left( \sum_{i=1}^{N} W_{d_{\mathcal{T}}} (\mu, \mu_i) \right).
\label{fig:fs_twb_chain}
\end{align}
Then, similar to the FS-TWB problem, the objective function can be rewritten as follows:
\begin{align}
    \mathbf{B} &= 
    \mathbf{w}_v \circ (\mathbf{I} - \mathbf{D}_{\text{par}})^{-1}, \\
    f(\mathbf{a}) &= \frac{1}{N} \sum_{i=1}^{N} \| \mathbf{B} \mathbf{a} - \mathbf{B} \mathbf{a}_i \|_1.
\end{align}
The formulation of Eq. \eqref{fig:fs_twb_chain} is same as the tree-Wasserstein barycenter on $\bm{V}$ \citep{le2020treewasserstein},
and their algorithm can solve the FS-TWB problem when $\mathcal{T}$ is a chain.
However, note that their algorithm can not be applied to the FS-TSWB problem when the set of trees $\{\mathcal{T}^{(t)}\}_{t=1}^{T}$ is a set of chains.

\subsection{FastPSD for Fixed Support Sliced Wasserstein Barycenter}
Because the depth of the chain is $O(|\mathbf{V}_{\text{leaf}}|)$, the number of non-zero elements in $\mathbf{B}$ is $O(|\mathbf{V}_{\text{leaf}}|^2)$.
Therefore, we require $O(|\mathbf{V}_{\text{leaf}}|^2)$ to compute $\mathbf{B}^\top \mathbf{z}^{(k)}$ and $\mathbf{B} \mathbf{a}^{(k)}$ in Algorithm \ref{alg:fast_psd}.
In this section, we show that, by utilizing the chain structure, 
$\mathbf{B} \mathbf{a}^{(k)}$ and $\mathbf{B}^\top \mathbf{z}^{(k)}$ can be computed in $O(|\mathbf{V}_{\text{leaf}}|)$.

Without a lack of generality, we can arrange the index of nodes such that $(v_{i+1}, v_i) \in \bm{E}$ for all $i \in [\![ |\bm{V}|-1 ]\!]$.
We then obtain the following:
\begin{align}
    \mathbf{D}_{\text{par}} &=
    \begin{pmatrix}
    0 & 1 & 0 & \ldots & 0 \\
      &  \ddots & \ddots &\ddots & \vdots \\
    \vdots &   & \ddots & \ddots & 0 \\
      &   &  & \ddots & 1 \\
    0 &   & \ldots &  & 0 \\
    \end{pmatrix}, \\
    (\mathbf{I} - \mathbf{D}_{\text{par}})^{-1} &= 
    \begin{pmatrix}
    1 & \ldots & 1 \\
      & \ddots & \vdots \\
    \text{{\Huge 0}}  & & 1  \\
    \end{pmatrix}, \\
    \mathbf{B} &= \mathbf{w}_v \circ (\mathbf{I} - \mathbf{D}_{\text{par}})^{-1} = \text{diag}(\mathbf{w}_v) (\mathbf{I} - \mathbf{D}_{\text{par}})^{-1},
\end{align}
where $\text{diag}(\mathbf{w}_v)$ denotes the diagonal matrix whose element in the $i$-th row and $i$-th column is $[\mathbf{w}_v]_i$.
Then, $\mathbf{B} \mathbf{a}^{(k)}$ can be computed as follows:
\begin{align}
    \mathbf{B} \mathbf{a}^{(k)} = \text{diag}(\mathbf{w}_v) \left( (\mathbf{I} - \mathbf{D}_{\text{par}})^{-1} \mathbf{a}^{(k)} \right).
\end{align}
Considering the property of $(\mathbf{I} - \mathbf{D}_{\text{par}})^{-1}$, 
we can compute $\mathbf{B} \mathbf{a}^{(k)}$ using Algorithm \ref{alg:fast_Ba},
whose time complexity is $O(|\bm{V}_{\text{leaf}}|)$.
\begin{algorithm}[tb]
   \caption{Fast computation for $\mathbf{B} \mathbf{a}^{(k)}$}
   \label{alg:fast_Ba}
\begin{algorithmic}[1]
   \STATE {\bfseries Input:}  $\mathbf{w}_v, \mathbf{a}^{(k)}$
   \STATE {\bfseries Output:} $\mathbf{B} \mathbf{a}^{(k)}$
   \STATE ${\mathbf{b}^\prime}^{(k)} \leftarrow \mathbf{0}_{|\bm{V}_{\text{leaf}}|}$
   \STATE $[{\mathbf{b}^\prime}^{(k)}]_{|\bm{V}_{\text{leaf}}|} \leftarrow [\mathbf{a}^{(k)}]_{|\bm{V}_{\text{leaf}}|}$
   \FOR{$i = |\bm{V}_{\text{leaf}}|-1, |\bm{V}_{\text{leaf}}|-2, \ldots, 1$}
   \STATE $[{\mathbf{b}^\prime}^{(k)}]_i \leftarrow [{\mathbf{b}^\prime}^{(k)}]_{i+1} + [\mathbf{a}^{(k)}]_i$
   \ENDFOR
   \STATE {\bfseries return} $\text{diag}(\mathbf{w}_v) {\mathbf{b}^\prime}^{(k)}$
\end{algorithmic}
\end{algorithm}

Next, similar to Algorithm \ref{alg:fast_Ba}, 
we show that $\mathbf{B}^\top \mathbf{z}^{(k)}$ can be computed in $O(|\bm{V}_{\text{leaf}}|)$.
Here, $\mathbf{B}^\top \mathbf{z}^{(k)}$ is computed as follows:
\begin{align}
    \mathbf{B}^\top \mathbf{z}^{(k)} = {(\mathbf{I} - \mathbf{D}_{\text{par}})^{-1}}^\top \left( \text{diag}(\mathbf{w}_v) \mathbf{z}^{(k)} \right).
\end{align}
Considering the property of $(\mathbf{I} - \mathbf{D}_{\text{par}})^{-1}$, 
we can compute $\mathbf{B}^\top \mathbf{z}^{(k)}$ using Algorithm \ref{alg:fast_Btz},
whose time complexity is $O(|\bm{V}_{\text{leaf}}|)$.
\begin{algorithm}[tb]
   \caption{Fast computation for $\mathbf{B}^\top \mathbf{z}^{(k)}$}
   \label{alg:fast_Btz}
   \begin{algorithmic}[1]
   \STATE {\bfseries Input:}  $\mathbf{w}_v, \mathbf{a}^{(k)}$
   \STATE {\bfseries Output:} ${\mathbf{B}}^\top  \mathbf{z}^{(k)}$
   \STATE ${\mathbf{z}^{\prime}}^{(k)} \leftarrow \text{diag}(\mathbf{w}_v) \mathbf{z}^{(k)}$
   \STATE ${\mathbf{g}^{\prime}}^{(k)} \leftarrow \mathbf{0}_{|\bm{V}_{\text{leaf}}|}$
   \STATE $[{\mathbf{g}^{\prime}}^{(k)}]_1 \leftarrow [{\mathbf{z}^{\prime}}{(k)}]_{1}$
   \FOR{$i = 2, 3, \ldots, |\bm{V}_{\text{leaf}}|$}
   \STATE $[{\mathbf{g}^{\prime}}^{(k)}]_i \leftarrow [\mathbf{b}^{(k)}]_{i-1} + [{\mathbf{z}^{\prime}}^{(k)}]_i$
   \ENDFOR
   \STATE {\bfseries return} ${\mathbf{g}^{\prime}}^{(k)}$
\end{algorithmic}
\end{algorithm}
In summary, 
when $\mathcal{T}$ is a chain,
the time complexity per iteration of the FastPSD can be reduced to $O(|\bm{V}_{\text{leaf}}| (\log(|\bm{V}_{\text{leaf}}|) + \log(N)))$
by using Algorithm \ref{alg:fast_Ba} and \ref{alg:fast_Btz}.

Similar to the discussion in Sec. \ref{sec:fs-tswb},
the FastPSD can be naturally extended to solve the FS-TSWB problem when the set of trees $\{ \mathcal{T}^{(t)} \}_{t=1}^{T}$ is the set of chains. 
(i.e., the fixed support sliced Wasserstein barycenter).
Then, using Algorithms \ref{alg:fast_Ba} and \ref{alg:fast_Btz}, 
the time complexity for each iteration of the FastPSD can be reduced to $O(T |\bm{V}_{\text{leaf}}| (\log(|\bm{V}_{\text{leaf}}|) + \log(N)))$.

\section{Additional Analyses of Time Consumption}
\begin{figure*}[h!]
\center
\subfigure[AMAZON]{
\includegraphics[width=0.3\textwidth]{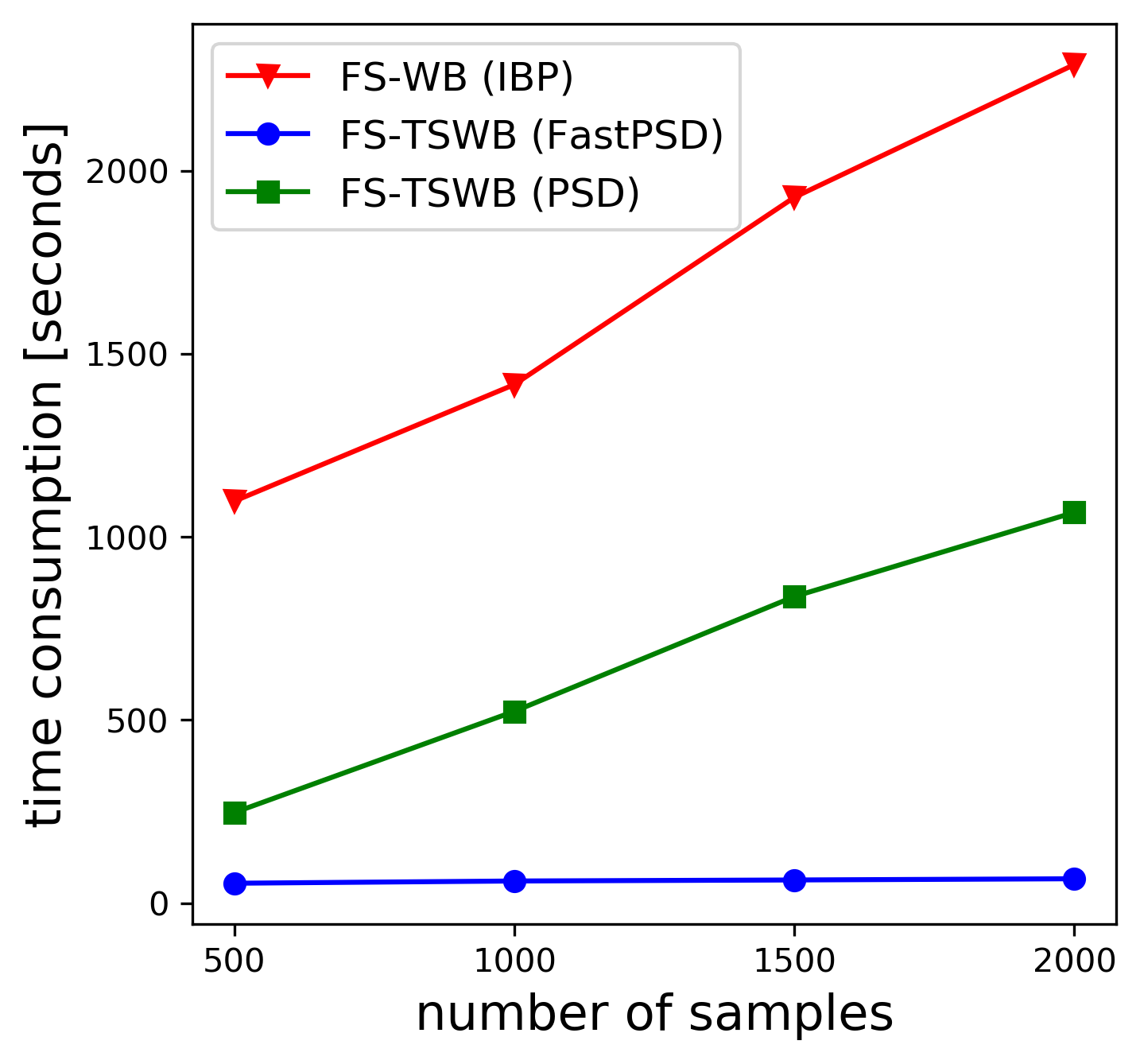}
}
\subfigure[AGNews]{
\includegraphics[width=0.3\textwidth]{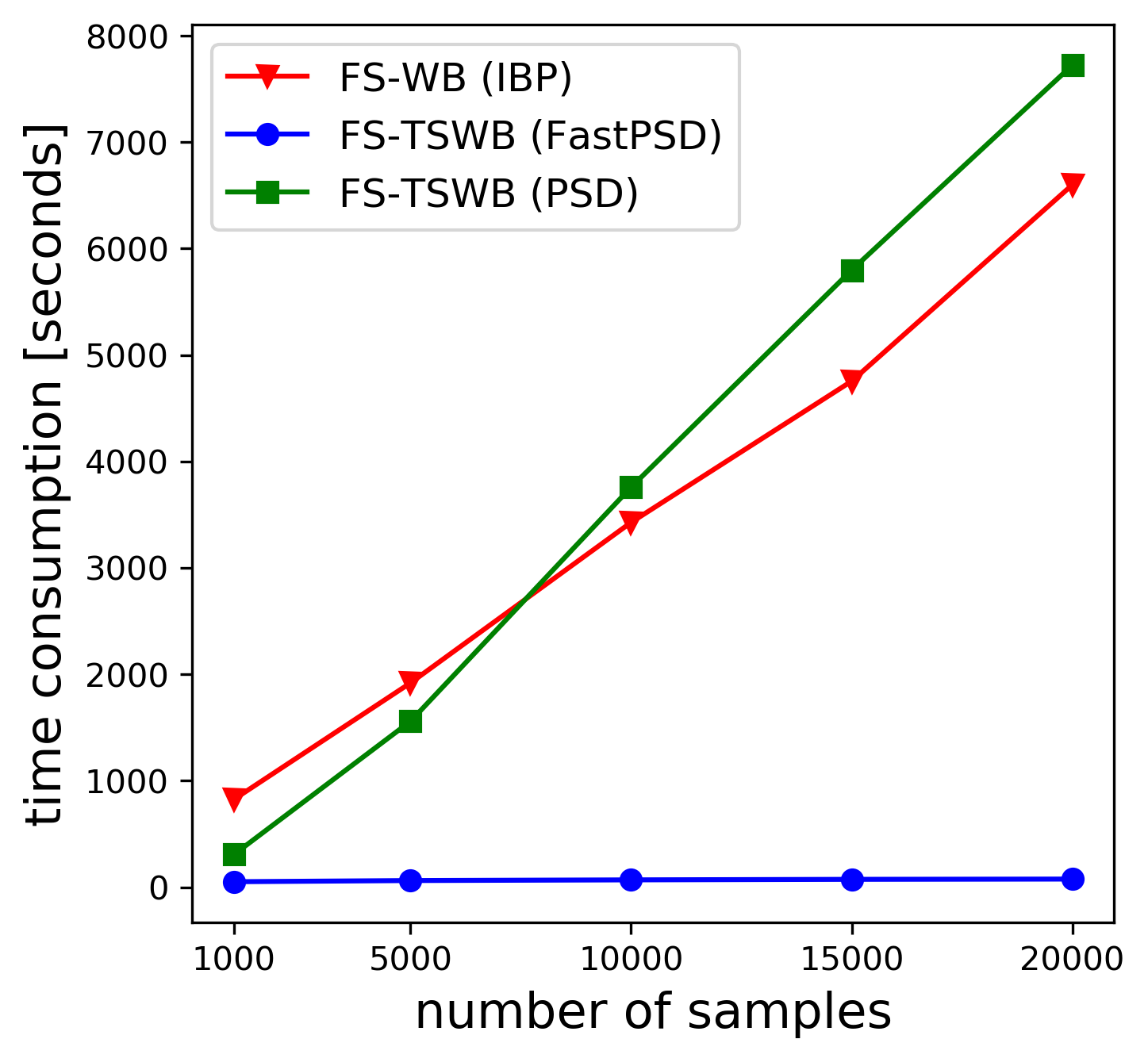}
}
\vskip -0.15 in
\caption{Time consumption when varying the number of samples.}
\label{fig:additional_time}
\end{figure*}

\begin{figure*}[h!]
\center
\subfigure{
\includegraphics[width=0.3\textwidth]{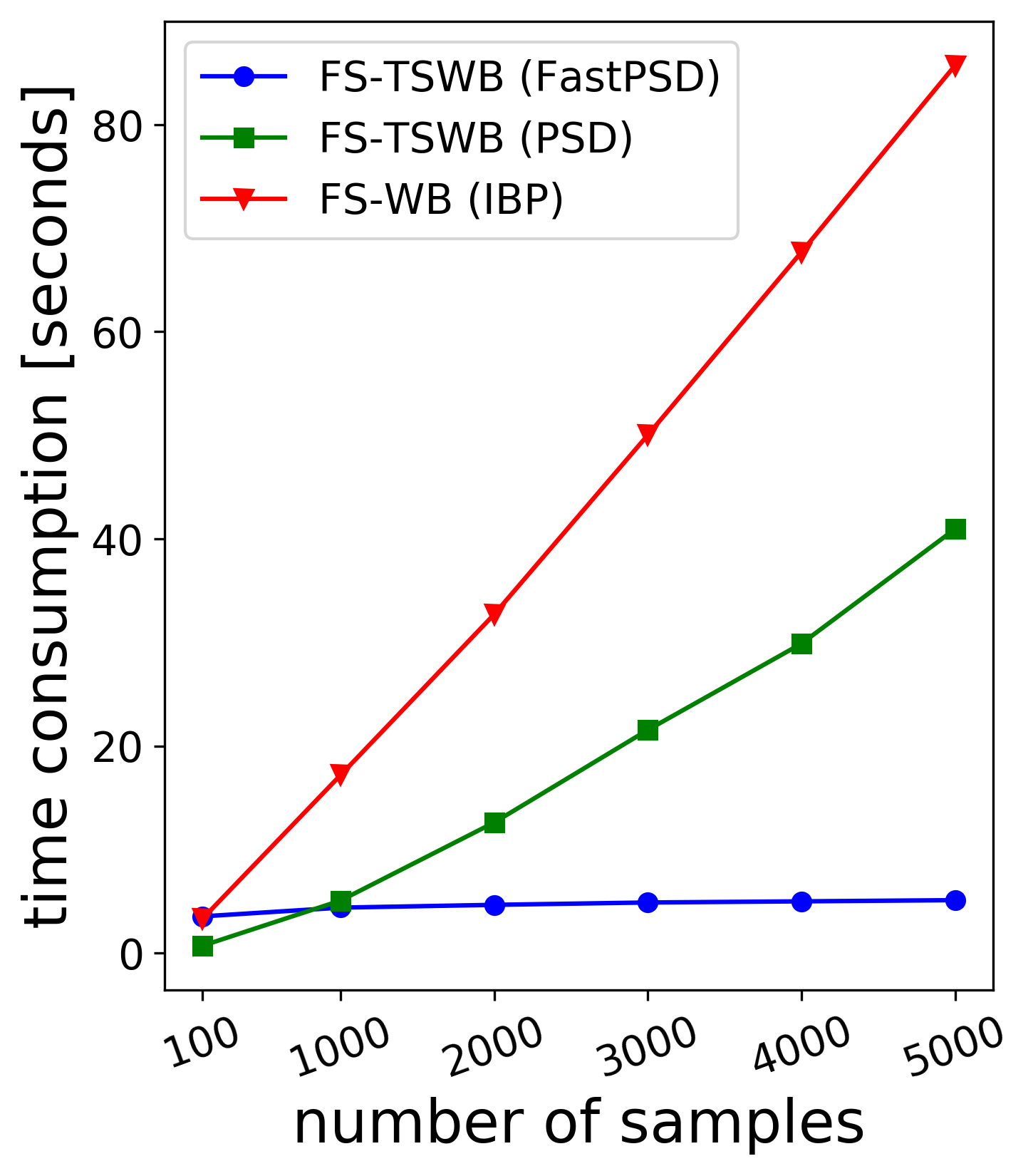}
}
\subfigure{
\includegraphics[width=0.31\textwidth]{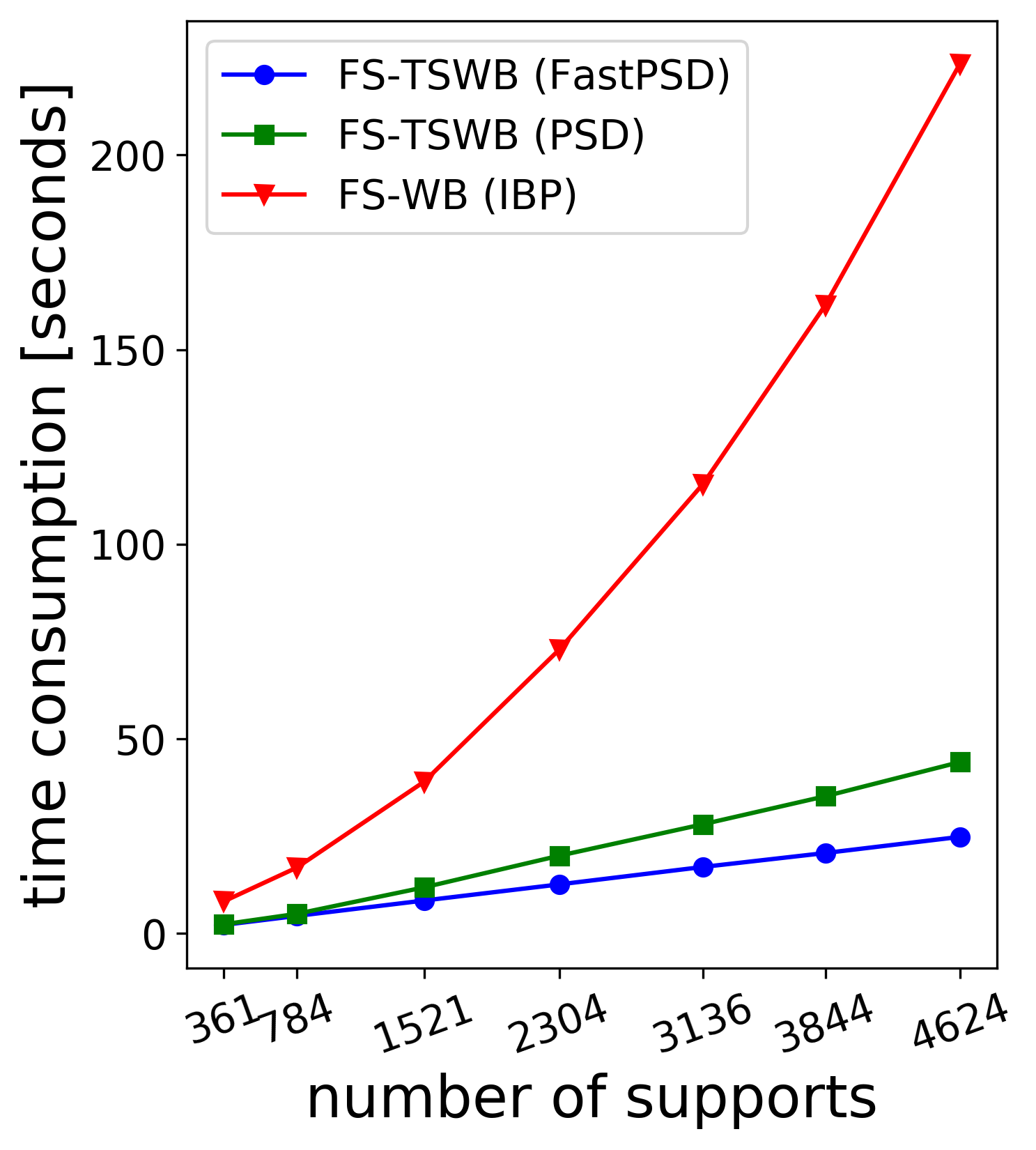}
}
\vskip -0.15 in
\caption{Time consumption when varying the number of samples and the number of supports on FashionMNIST. The number of tree is set to one. When the number of supports is varied, the number of samples is set to 1000. The results are averages for all categories.}
\label{fig:fashion_speed}
\end{figure*}

\begin{figure*}[h!]
\center
\subfigure{
\includegraphics[width=0.3\textwidth]{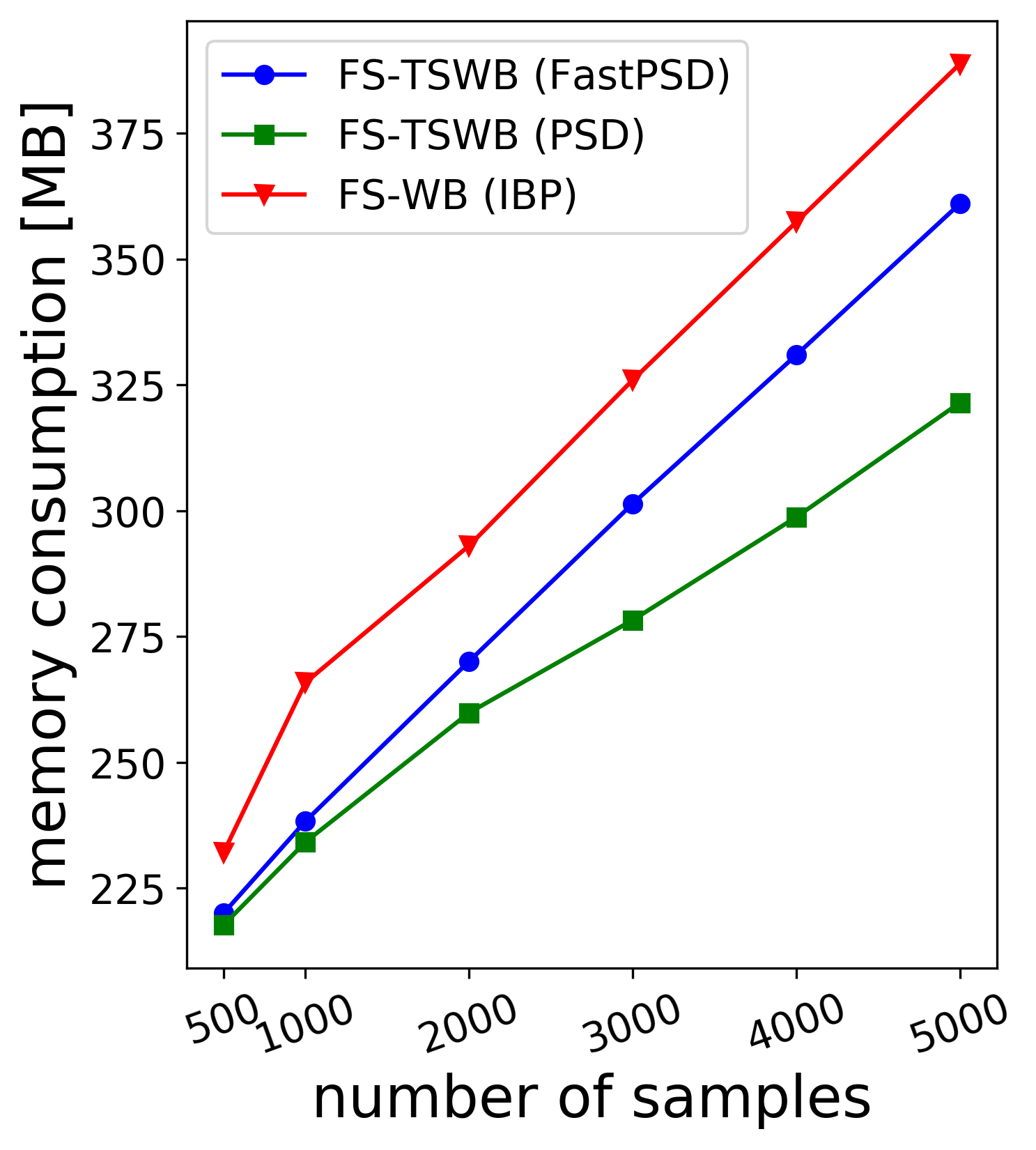}
}
\subfigure{
\includegraphics[width=0.3\textwidth]{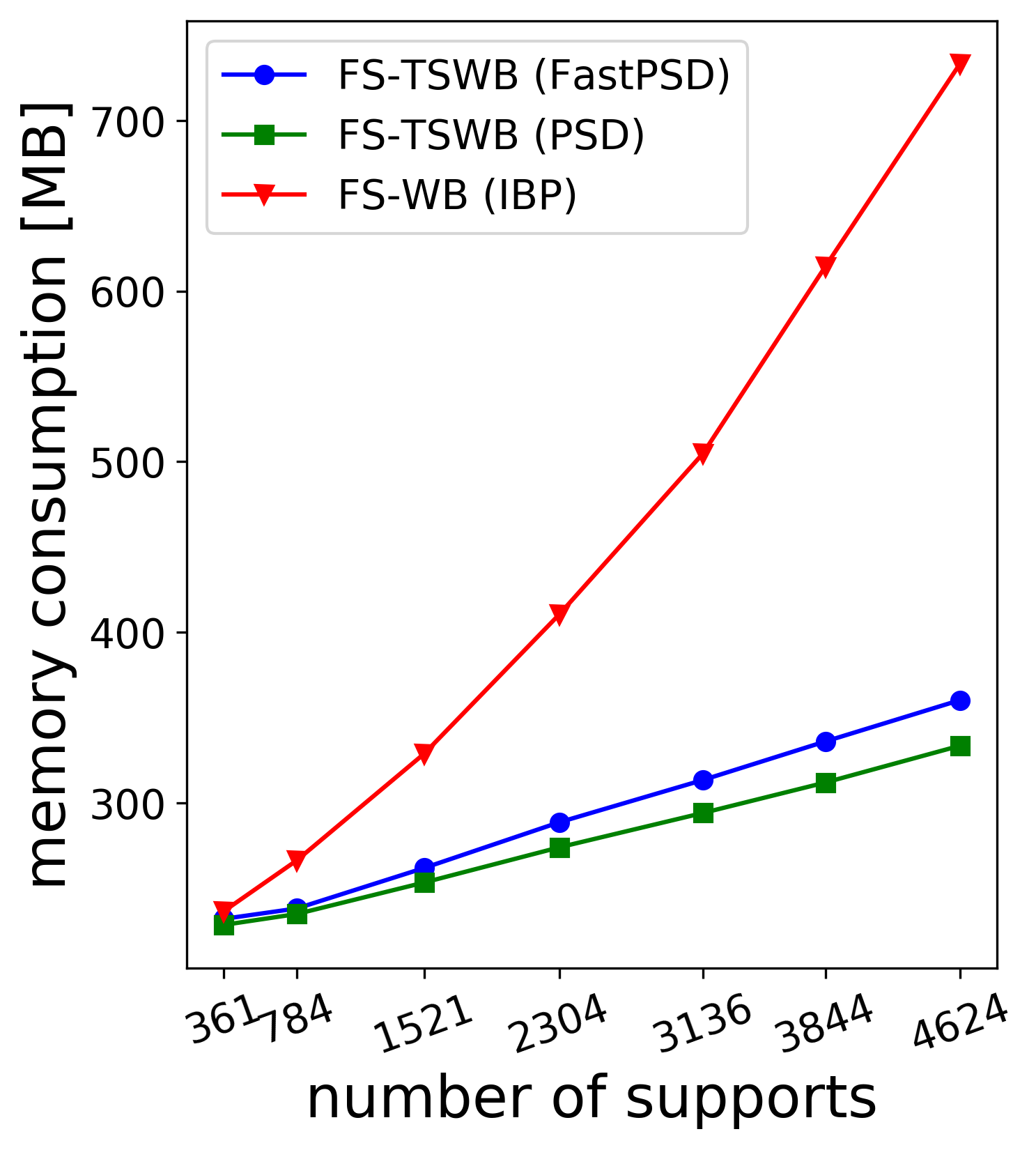}
}
\vskip -0.15 in
\caption{Memory consumption when varying the number of samples and the number of supports on FashionMNIST. The number of tree is set to one. When the number of supports is varied, the number of samples is set to 1000. The results are averages for all categories.}
\label{fig:fashion_memory}
\end{figure*}

Fig. \ref{fig:additional_time} shows the time consumption varying the number of samples on AMAZON and AGNews.
Figs. \ref{fig:fashion_speed} and \ref{fig:fashion_memory} show the time consumption and the memory consumption on FashionMNIST.

\newpage
\section{Visualization of Barycenters}
\label{sec:other_visualization}
\begin{figure}[h]
\centering
\includegraphics[width=\hsize]{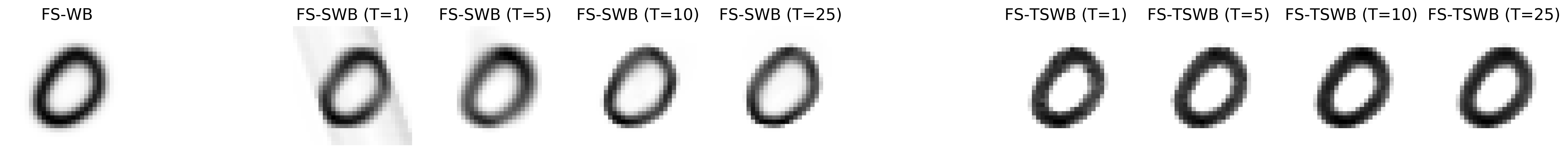}
\includegraphics[width=\hsize]{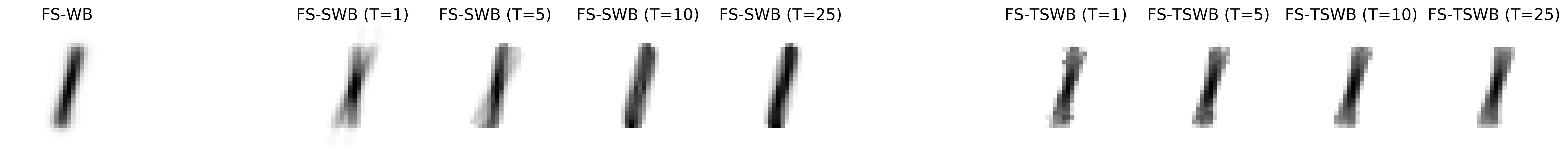}
\includegraphics[width=\hsize]{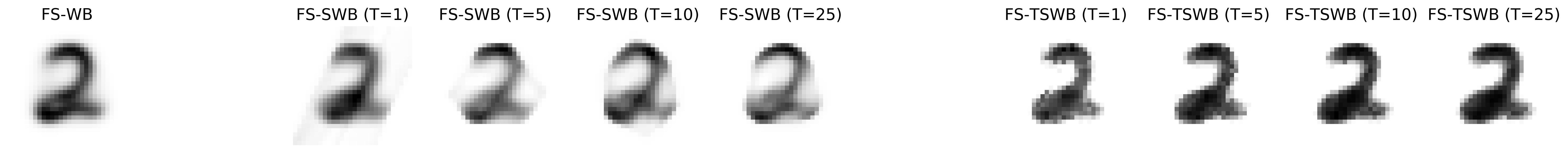}
\includegraphics[width=\hsize]{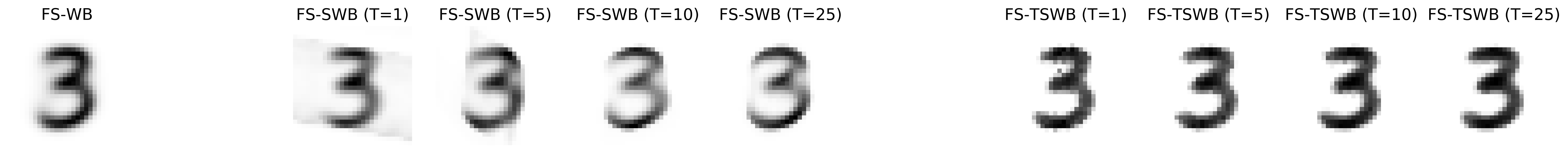}
\includegraphics[width=\hsize]{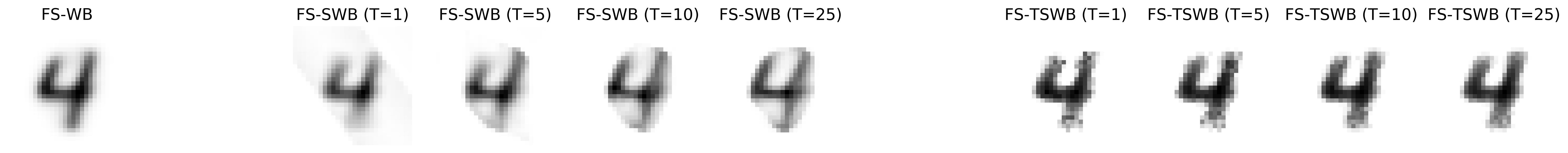}
\includegraphics[width=\hsize]{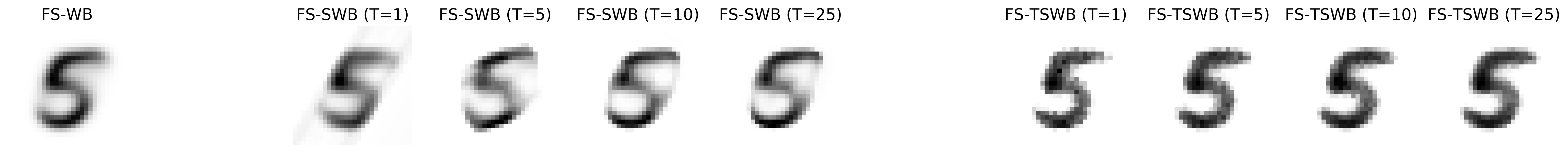}
\includegraphics[width=\hsize]{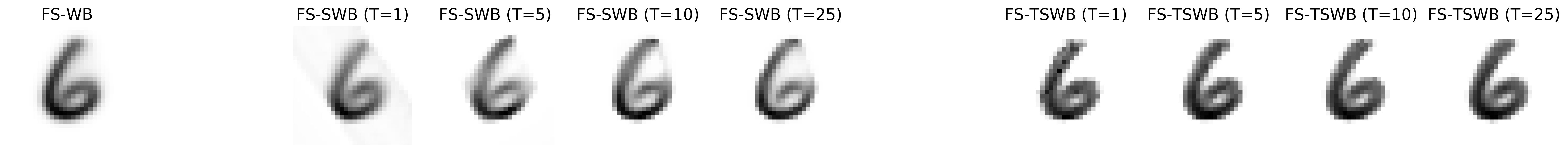}
\includegraphics[width=\hsize]{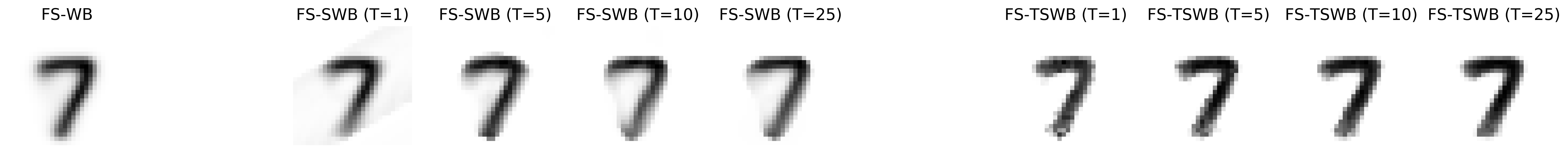}
\includegraphics[width=\hsize]{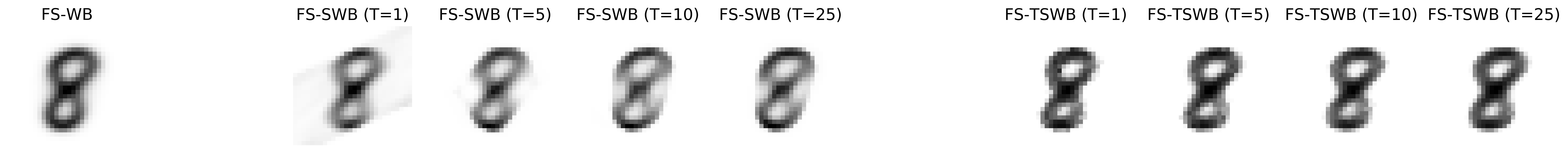}
\includegraphics[width=\hsize]{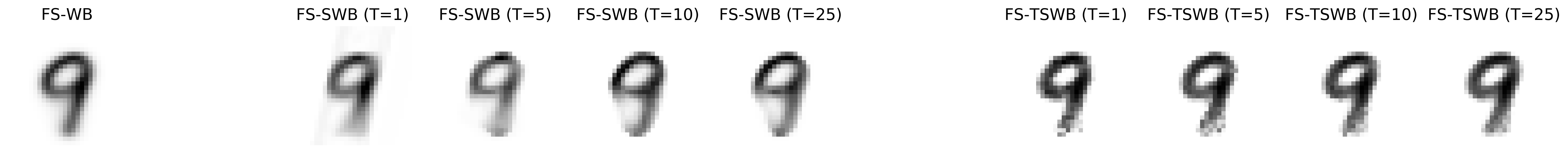}
\caption{Visualization of the FS-WB, the FS-SWB, and the FS-TSWB on MNIST.}
\end{figure}

\newpage
\begin{figure}[h]
    \centering
    \subfigure[FS-WB]{
        \includegraphics[width=0.75\hsize]{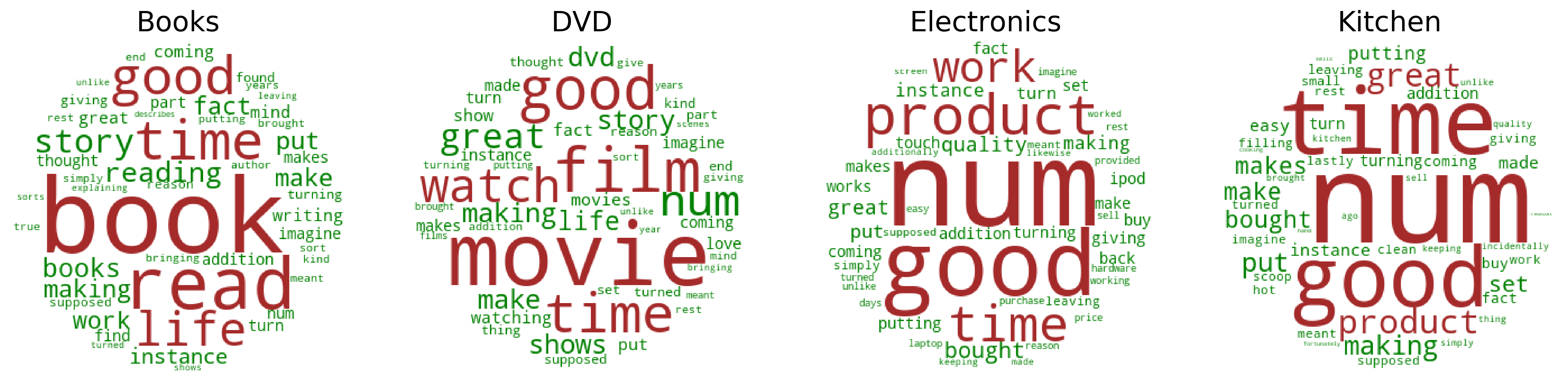}
    }
    \subfigure[FS-TSWB ($T=1$)]{
        \includegraphics[width=0.75\hsize]{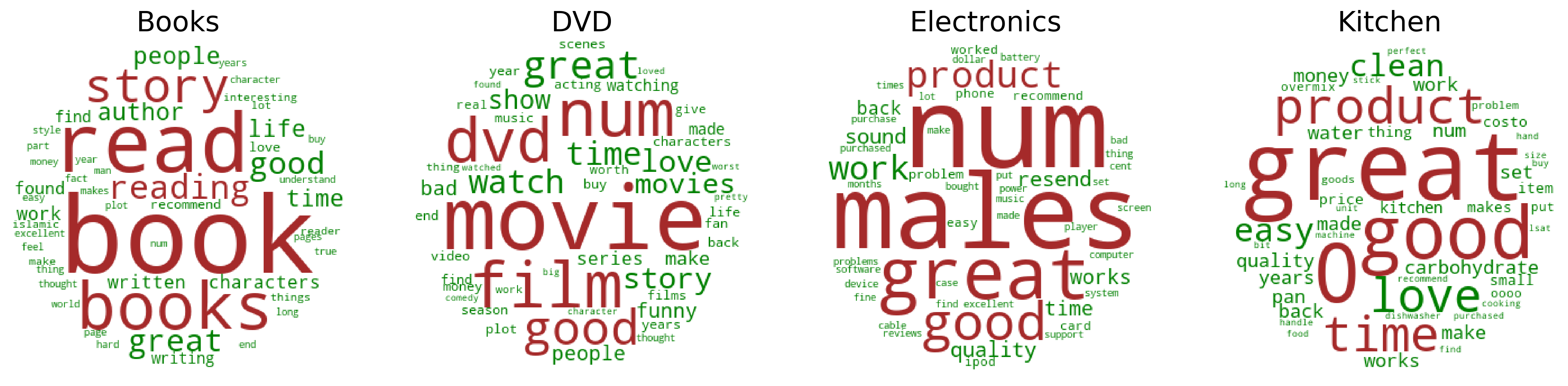}
    }
    \subfigure[FS-TSWB ($T=25$)]{
        \includegraphics[width=0.75\hsize]{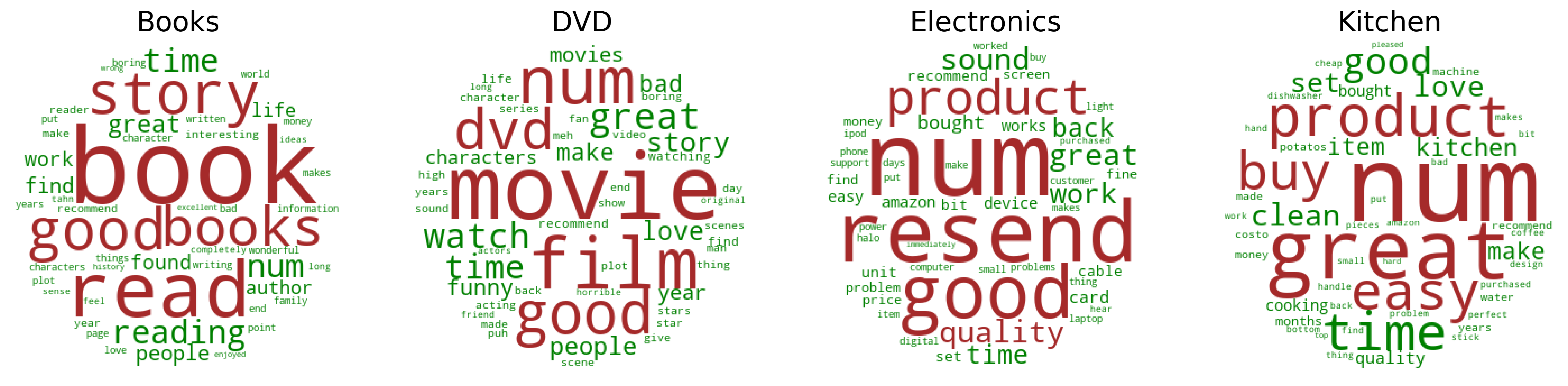}
    }
    \subfigure[FS-SWB ($T=1$)]{
        \includegraphics[width=0.75\hsize]{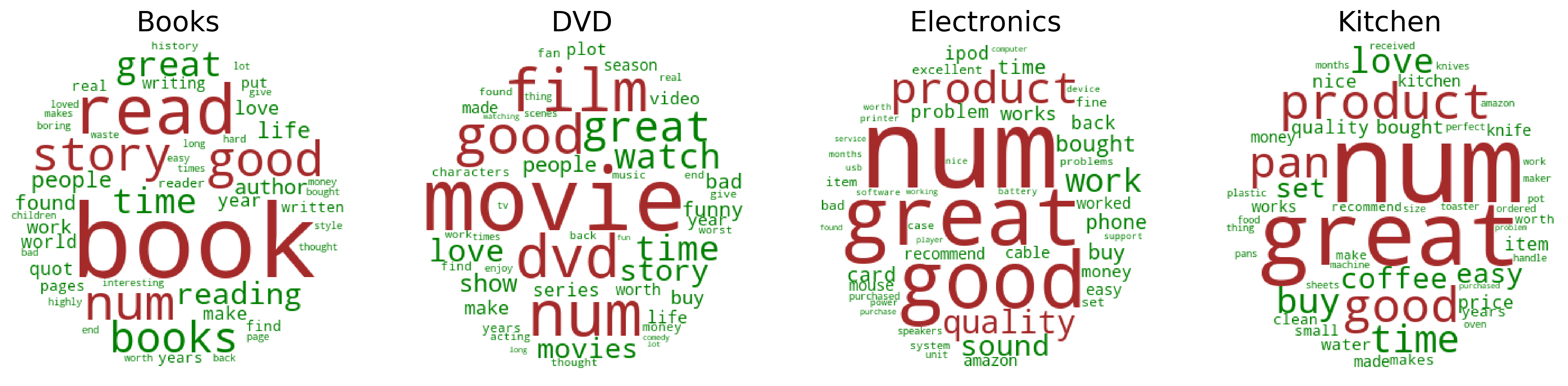}
    }
    \subfigure[FS-SWB ($T=25$)]{
        \includegraphics[width=0.75\hsize]{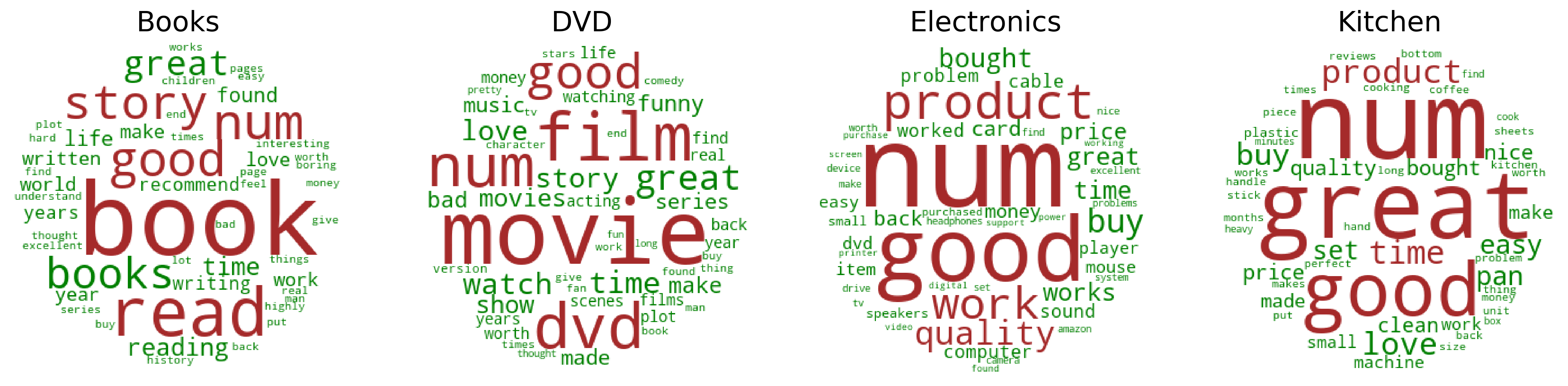}
    }
    \caption{Visualization of the FS-WB, the FS-SWB, and the FS-TSWB on AMAZON.}
\end{figure}

\newpage
\begin{figure}[h]
    \centering
    \subfigure[FS-WB]{
        \includegraphics[width=0.75\hsize]{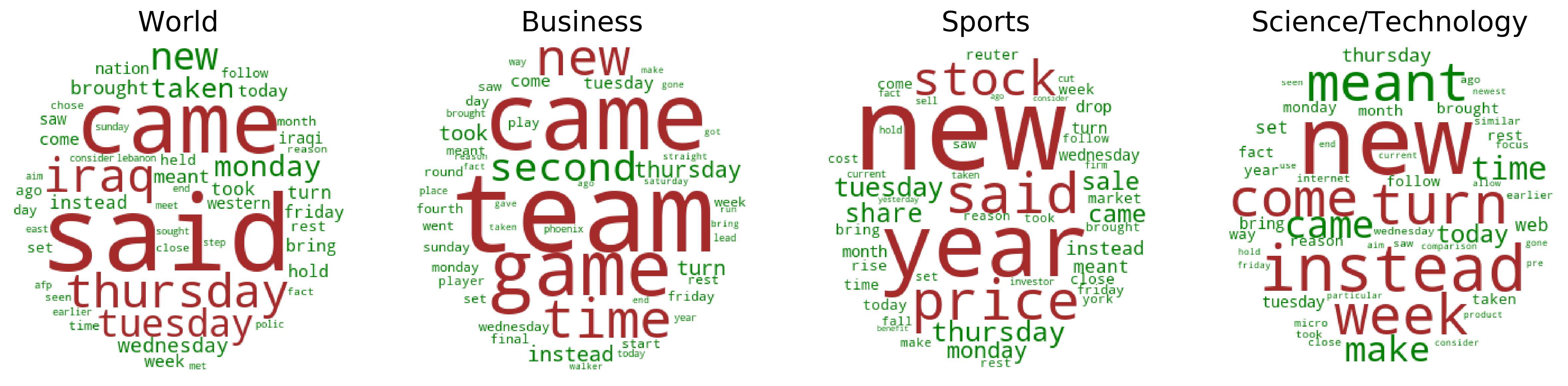}
    }
    \subfigure[FS-TSWB ($T=1$)]{
        \includegraphics[width=0.75\hsize]{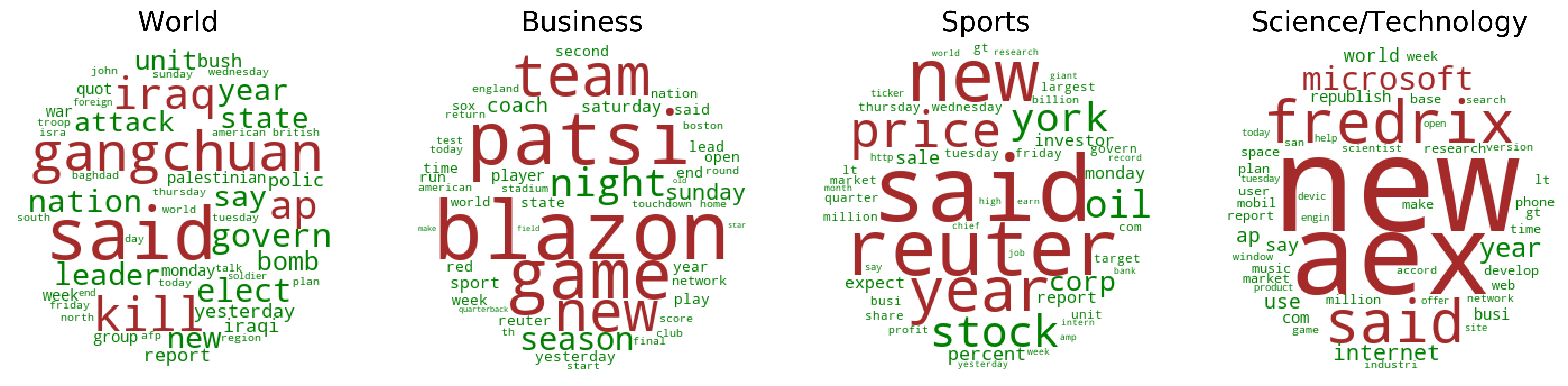}
    }
    \subfigure[FS-TSWB ($T=25$)]{
        \includegraphics[width=0.75\hsize]{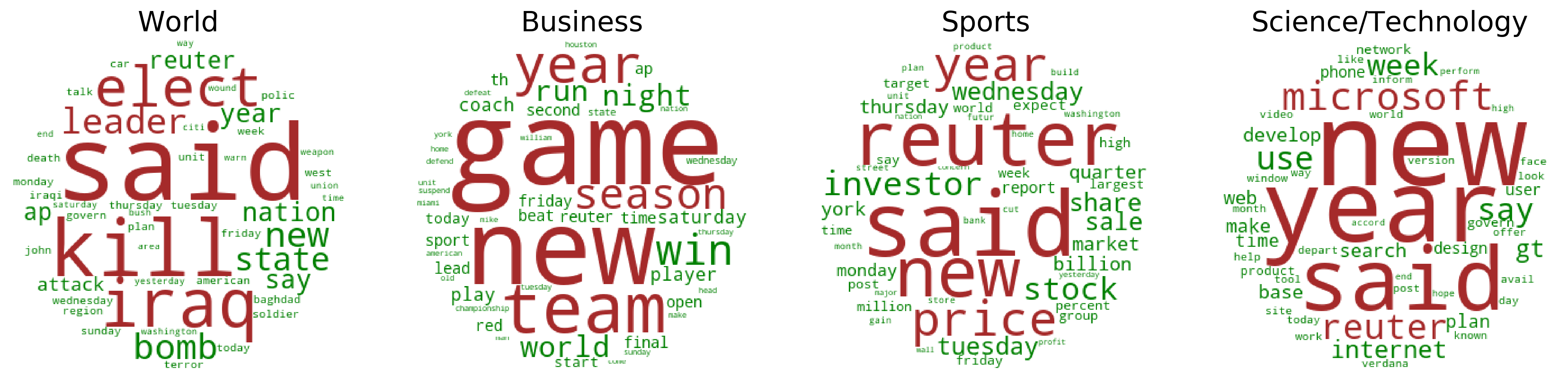}
    }
    \subfigure[FS-SWB ($T=1$)]{
        \includegraphics[width=0.75\hsize]{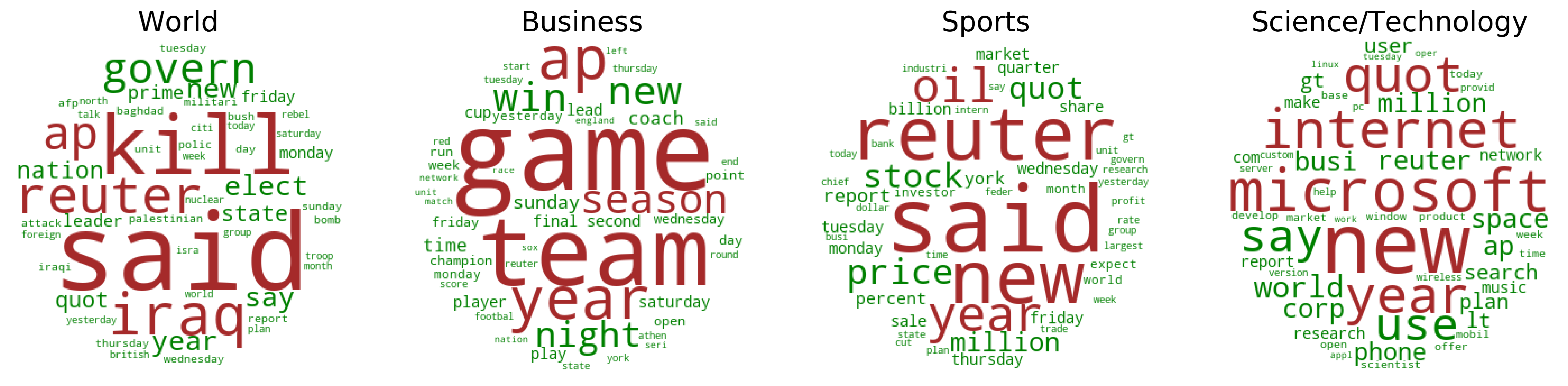}
    }
    \subfigure[FS-SWB ($T=25$)]{
        \includegraphics[width=0.75\hsize]{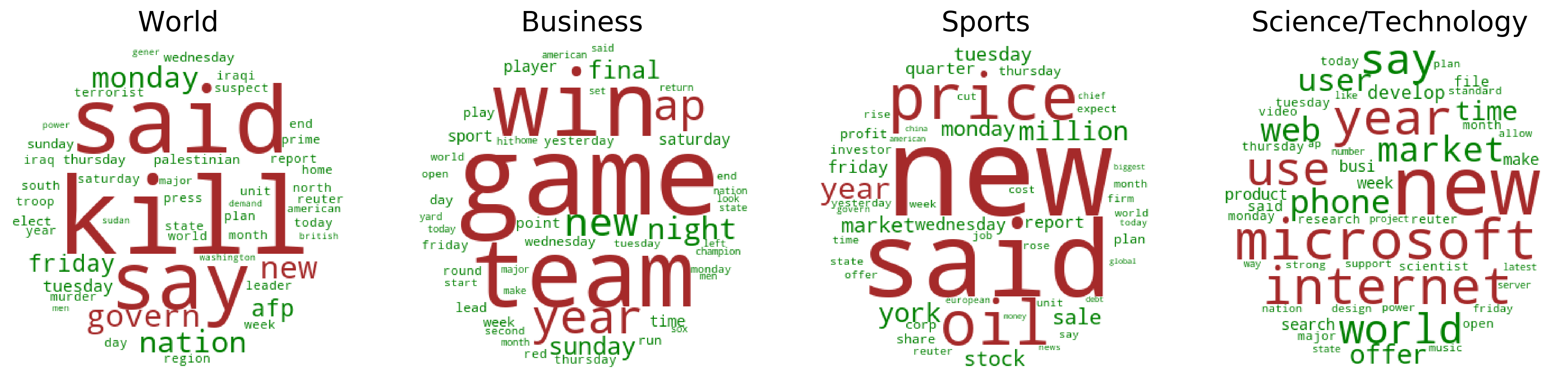}
    }
    \caption{Visualization of the FS-WB, the FS-SWB, and the FS-TSWB on AGNews.}
\end{figure}

\end{document}